\documentclass[letterpaper, 10 pt, conference]{ieeeconf}  

\IEEEoverridecommandlockouts                              

\usepackage[inkscapelatex=false]{svg}
\usepackage{amsmath}
\usepackage{amssymb}
\usepackage{graphicx}
\usepackage{textcomp}
\usepackage{xcolor}
\usepackage{pgfplots}
\usepackage{subfigure}
\usepgfplotslibrary{groupplots}
\usepackage{tikz}
\usepackage{tikzscale}
\usepackage{hyperref}
\usepackage{cleveref}
\usepackage{float}
\usepackage{multirow}
\usepackage{caption}
\usepackage{algorithm}
\usepackage{algpseudocode} 
\usepackage{siunitx}
\usepackage{comment}
\usepackage{booktabs}
\usepackage{alphalph}
\usepackage{todonotes}
\usepackage{pifont}

\crefname{figure}{Fig.}{Fig.}
\Crefname{figure}{Figure}{Figures}
\crefname{equation}{}{}
\Crefname{equation}{Equation}{Equations}

\newcommand{\ie}{\textit{i}.\textit{e}., }
\newcommand{\eg}{\textit{e}.\textit{g}., }

\title{\LARGE \bf
AssemblyComplete: 3D Combinatorial Construction\\with Deep Reinforcement Learning
}

\author{Alan Chen$^{1}$, Changliu Liu$^{2}$
\thanks{$^{1}$Alan Chen is with Westlake High School, 
        Austin, TX, 78746, USA. 
        {\tt\small alanjiach@gmail.com}
}
\thanks{$^{2}$Changliu Liu is with the Robotics Institute,
	Carnegie Mellon University,
	Pittsburgh, PA, 15213, USA.
        {\tt\small cliu6@andrew.cmu.edu}
}}

\begin{document}

\maketitle
\thispagestyle{empty}
\pagestyle{empty}

\begin{abstract}
A critical goal in robotics and autonomy is to teach robots to adapt to real-world collaborative tasks, particularly in automatic assembly. 
The ability of a robot to understand the original intent of an incomplete assembly and complete missing features without human instruction is valuable but challenging.  
This paper introduces 3D combinatorial assembly completion, which is demonstrated using combinatorial unit primitives (\ie Lego bricks). 
Combinatorial assembly is challenging due to the possible assembly combinations and complex physical constraints (\eg no brick collisions, structure stability, inventory constraints, etc.). 
To address these challenges, we propose a two-part deep reinforcement learning (DRL) framework that tackles teaching the robot to understand the objective of an incomplete assembly and learning a construction policy to complete the assembly.
The robot queries a stable object library to facilitate assembly inference and guide learning.
In addition to the robot policy, an action mask is developed to rule out invalid actions that violate physical constraints for object-orientated construction. 
We demonstrate the proposed framework's feasibility and robustness in a variety of assembly scenarios in which the robot satisfies real-life assembly with respect to both solution and runtime quality. 
Furthermore, results demonstrate that the proposed framework effectively infers and assembles incomplete structures for unseen and unique object types.
\end{abstract}

\section{Introduction}

Recent advancements in robotics and computer vision research have enabled robots to learn valuable skills for collaborative manufacturing tasks, such as learning from human demonstration \cite{kyrarini2019robot, liu2023salfd}, collaborative robots \cite{serocs, ChenRUi}, and assembly sequence planning \cite{ORASP, GraphASP, MASEHIAN2021102180}, which are important in achieving a fully autonomous assembly line. 

3D assembly completion remains relatively unexplored, yet it is essential for robot autonomy and collaboration. 
If given the capability to infer and finish incomplete 3D assemblies, robots can master human-robot collaboration.  
For instance, consider a scenario where a human is assembling a chair but cannot finish it due to limitations. 
In this case, a robot could take over, analyze the incomplete assembly, and finish it without human guidance. 
This skill extends to enhancing the efficiency of multi-robot systems as well, as it allows robots to be more flexible and access complicated skills like streamlined collaboration with other robots.

In this paper, we introduce the problem of 3D combinatorial assembly completion. 
This task is challenging as the robot's goal is to finish the incomplete assembly without prior knowledge about the structure's end goal. 
Moreover, incomplete assemblies often lack necessary information, such as entire missing features, and the robot may fail to identify and assemble those features.
Robots generally cannot grasp real-world physics and may overlook adding key structural elements that enforce and connect sub-components to support each other. 
As a result, the structure may collapse later on.
Thus, we teach the robot to infer from the incomplete assembly what the previous builder intended to build using a stable reference structure stored within an object library. 


\begin{figure*}
    \centering
    \includegraphics[width=\linewidth]{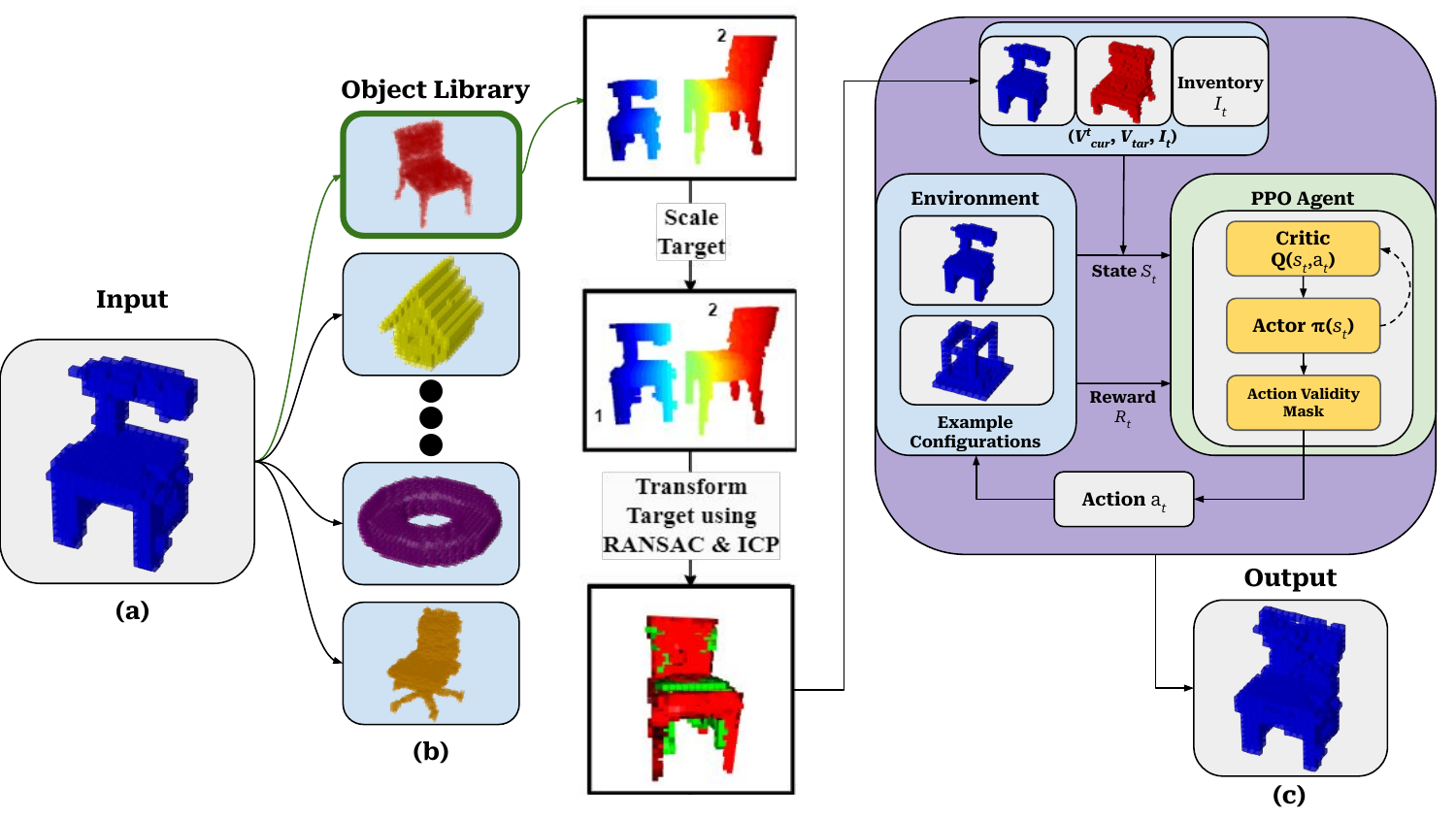}
    \vspace{-10pt}
    \caption{\footnotesize 
    Our method is designed to: (a) Represent an incomplete structure as a voxel and point cloud. (b) Incorporate a library of real-world objects for inference and matching. In this step, target scaling with the best-matched complete object from the library and 3D point cloud registration are performed. 1 represents the incomplete/source point cloud. 2 represents the completed/target point cloud. (c) Train a DRL agent to use heuristics and satisfy multiple constraints for efficient sequential decision-making. Here, given the current state $s_t$ (incomplete structure being built, reference structure, and inventory), the PPO agent learns to choose an optimal action and improve its policy and value function using a reward. 
    \label{fig:overall_pipeline}}
    \vspace{-15pt}
\end{figure*}

Traditional assembly completion works \cite{Zhang_2022, AssemblyCompletion} explore incomplete part assembly where the goal is to assemble an object using whole unique parts (\ie chair arm, leg, backrest).
In contrast, combinatorial assembly completion uses unit primitives (\ie Lego bricks, etc.) to model real-world sequential assembly. 
Legos are a popular model used in assembly planning as demonstrated by \cite{Kim2014SurveyOA, KimJ2020arxiv, ChungH2021neurips} as Legos allow for customizability, creativity, and showcase additional challenges over traditional part assembly. 

This is because, unlike traditional part assembly, combinatorial assembly has to consider additional physical constraints such as a limited resource inventory, a significantly larger number of combinations, structural integrity, and more.
Legos also provide unique challenges, such as their ability to latch and hook themselves on top or below another brick, as shown in \cref{fig:constraints}.
Furthermore, combinatorial assembly can directly handle part assembly as a unit primitive can be defined as a larger pre-constructed part (\eg chair arm, leg, backrest).

In summary, this paper studies the 3D assembly completion task by proposing a framework that understands the intention of the incomplete structure and generates sequential instruction to completion while satisfying various constraints. 
Our main contributions are:
\begin{enumerate}
    \item We develop a novel application of 3D point cloud data (PCD) that allows feature matching within an object library that overcomes scale and rotational invariances to enable robots to infer autonomously, refer to parts (a)\&(b) in \cref{fig:overall_pipeline}.
    \item We implement a DRL agent with Proximal Policy Optimization (PPO) to address incomplete combinatorial construction and optimize similarity to reference assembly, refer to part (c) in \cref{fig:overall_pipeline}.
    \item We enable the agent to realize real-world constraints (\eg structural feasibility, inventory, etc.) through a heuristic-based action mask for policy training.
    
    Contributions 1 and 3 do not require any training because 1 uses non-data point cloud algorithms, and 3 uses a heuristic-based action mask.
\end{enumerate}

\section{Related Works}
\subsection{3D Shape Completion}
3D shape completion is a deeply studied field in computer vision and robotics as it aims to reconstruct incomplete objects by filling in missing features. 
However, it is challenging to process and understand 3D geometric shapes. 
Early works \cite{10.1111/cgf.12010, :10.2312/SGP/SGP05/023-032} relied heavily on 3D geometric methods like symmetry and template matching but were limited when dealing with complex shapes and noisy data. 
Recent advancements have introduced deep-learning and generative-based techniques for improved accuracy of reconstructed shapes. 
\cite{CPVRPointSetGen} introduces a point set generation network that outperformed previous state-of-the-art methods for 3D reconstruction. 
Generative methods \cite{wu2020multimodalshapecompletionconditional, zheng2022sdfstyleganimplicitsdfbasedstylegan, mittal2023autosdfshapepriors3d}, like generative adversarial networks and variational auto-encoders, are favored in shape reconstruction.
Recent works have also explored a diffusion-based approach \cite{chengetal, chu2023diffcompletediffusionbasedgenerative3d}, which progressively refines shape details and showcases the effectiveness of text-to-image models.
In relation to these works, we use a similar 3D PCD representation to model features and geometry.
However, our work differs as we aim to reconstruct 3D shapes in a dynamic environment where we consider real-world assembly and physical constraints, as shown in \cref{fig:constraints}.

\subsection{3D Assembly Completion}

3D assembly completion aims to reconstruct objects while considering constraints that model manufacturing and assembly line applications. 
Current assembly completion works tackle in-process part assembly.
Similar to shape completion, generative and encoder methods are favored solutions for the 3D part assembly task. 
For example, Wu et al. \cite{wu2020pqnetgenerativeseq2seqnetwork} developed PQ-NET, a generative network for 3D shapes via sequential part assembly. 
Zhang et al. \cite{9813581} focus on 3D part assembly generation using an instance-encoded transformer. 
Wang et al. \cite{AssemblyCompletion} proposed FiT, a transformer approach to complete incomplete part assemblies. 
However, these works tackle part assembly where the action space is very limited.
In comparison, we formulate 3D assembly completion as a combinatorial assembly task where unit primitives can create up to billions of possible sequences in a simple 48x48 workspace. 
\cite{GraphTransformerASP} uses graph transformers, Legos, and an input target structure graph to predict a sequence for in-process assembly but fails in medium-length sequences because of the vast combinations. 
DRL proves to be more effective in combinatorial works \cite{ChungH2021neurips, ASPCoRL}.
However, \cite{ChungH2021neurips} relies on a 2D image/graph and demonstrates surface-level assemblies.
Previous work \cite{ASPCoRL} tackles combinatorial assembly sequence planning (ASP) which is much different from combinatorial assembly completion. 
The goal of the previous work is to sequentially assemble a stable, pre-defined target voxel with a given inventory.
In this paper, the goal is to take an in-process assembly, understand what is being built, and finish it accordingly.
Since assembly completion is fairly similar, we create a similar definition to \cite{ ASPCoRL} with the inventory constraint and action definition.
However, because of the nature of assembly completion where no exact target voxel is given and a matched object will never be the same as the incomplete assembly, the action space is significantly expanded, which proves problematic for computation time in the previous work's stability mask approach. 
Additionally, the operability mask approach cannot be considered as an incomplete assembly may already have a feature above constraining the action.
Therefore, we define entirely different rewards and action mask heuristics that can effectively tackle combinatorial assembly completion. 

\section{Problem Formulation}\label{sec:prob_formulation}
Given an incomplete 3D structure $\mathcal{V}_{cur}$, the overall goal is to complete the incomplete structure using $\mathcal{I}$, a dictionary that tracks an inventory of available bricks. 

First, we formulate intention inference as an optimization problem where the robot must \textit{infer by matching} the incomplete structure with the most similar object from an object library of complete stable assemblies to obtain $\mathcal{V}_{tar}$, a 3D reference $h$x$w$x$d$ binary matrix.

Then we formulate assembly completion as a DRL problem using the Markov Decision Process (MDP), denoted as $\mathcal{M} = (\mathcal{S}, \mathcal{A}, \mathcal{P}, \mathcal{R})$. In this MDP, the state representation is defined as $\mathcal{S}=(\mathcal{V}_{cur}, \mathcal{V}_{\text{tar}}, \mathcal{I})$, while $\mathcal{A}$ denotes the set of actions, each placing a brick of unique location, orientation, and size in a 3D space.
The transition function $\mathcal{P}: \mathcal{S} \times \mathcal{A} \rightarrow \mathcal{S}$ defines the next state given the current state and action, and the reward function $\mathcal{R}: \mathcal{S} \times \mathcal{A} \rightarrow \mathbb{R}$ provides a reward based on the agent's selected action. 
The discount factor $\gamma \in [0, 1)$ accounts for future rewards. 
The objective is for the agent to learn a policy $\pi(a_t | s_t)$ that maximizes the expected cumulative discounted reward $J(\pi) = \mathbb{E}[\sum_{t=0}^{T} \gamma^t r_t]$, where $r_t$ is the reward at timestep $t$.

\begin{figure}
\subfigure[Inventory Usage.]{\includegraphics[width=0.331\linewidth]{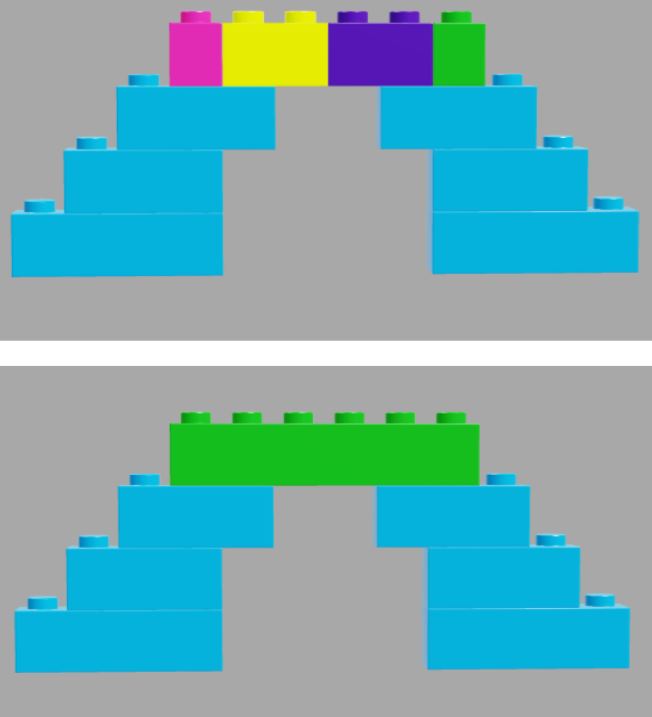}}\label{fig:inv_usage}\hfill
\subfigure[Hook Mechanism.]{\includegraphics[width=0.33\linewidth]{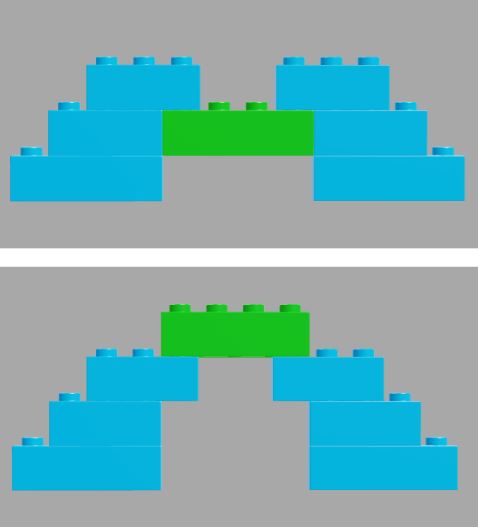}}\label{fig:hook}\hfill
\subfigure[Stability.]{\includegraphics[width=0.327\linewidth]{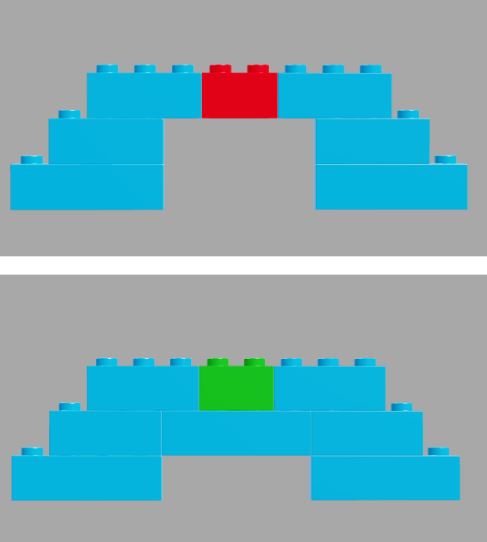}\label{fig:stable}}\hfill
\vspace{-5pt}
    \caption{\footnotesize Real-World Physical Constraints. \textit{Red}: invalid placement. \textit{Green}: valid placement.   
    \label{fig:constraints}}
    \vspace{-20pt}
\end{figure}

\section{Method}
The proposed framework in \cref{fig:overall_pipeline} consists of two major components: 1) inferring the intent of and matching the incomplete structure using 3D PCD and 2) completing the assembly using DRL.
First, we model the incomplete structure as PCD and apply feature registration and a similarity search on an object library to identify a matched assembly. 
We scale and transform the matched assembly to guide the incomplete assembly. 
Second, we train a DRL agent to understand the incomplete structure and sequentially assemble unit primitives under constraints to finish the structure with features similar to those of the matched assembly target.

\subsection{Inference and Matching Step}\label{infer&match}

In this section, we explain how we teach the robot to infer the incomplete assembly's end goal and process it to be used for DRL. 
We convert the incomplete assembly voxel into a 3D PCD and query against an object library to find the best match to determine the intent of the incomplete assembly. Refer to parts (a)\&(b) in \cref{fig:overall_pipeline}.


\subsubsection{Point Cloud Scaling}
To begin, we assume that the 3D incomplete assembly cannot be scaled or rotated, as it has already been physically constructed. 
In the real world, objects have distinct scales and rotations, making it impractical to constrain all structures to a preset configuration. 
Therefore, we adjust the target PCD to fit the incomplete assembly, focusing on object features for comparison.

We use Principal Component Analysis (PCA) to scale the target structure to fit the incomplete structure.
\begin{equation}\label{eq:pca_scaling}
\begin{split}
    \mathbf{P}_2^{\text{scaled}} &= \mathbf{P}_2 \cdot \max \left( \frac{\sqrt{\sigma_{1,1}}}{\sqrt{\sigma_{2,1}}}, \frac{\sqrt{\sigma_{1,2}}}{\sqrt{\sigma_{2,2}}}, \frac{\sqrt{\sigma_{1,3}}}{\sqrt{\sigma_{2,3}}} \right)
\end{split}
\end{equation}

Here, \(\sigma_{1,\textit{i}}\) and \(\sigma_{2,\textit{i}}\) represent the first and second structures' explained variance along the \textit{i}-th principal component, respectively.
The explained variance indicates how much of the data's variability is captured by each principal component.

\subsubsection{Point Cloud Registration}
After scaling the target PCD, we need to align the incomplete and target structures through their geometric features. 
However, this is challenging as the incomplete structures may have entirely missing features.
Therefore, feature descriptors are employed as local descriptors for the object's geometric features.
FPFH encodes information about neighbor nodes and direct pairs between query points, which is crucial for feature matching. 

For initial alignment, RANSAC is used to align the incomplete and target structures. 
Then, ICP is used to refine the alignment across multiple iterations, progressively reducing the distance threshold to achieve precision. 
The resulting transformation matrix is then applied to the target PCD and converted back to a voxel representation for the DRL to process. The 3D point cloud registration methods are implemented using the Open3D library \cite{zhou2018open3dmodernlibrary3d}.

\subsubsection{Similarity Matching}
Then, we compare stable and complete assemblies from an object library with the incomplete assembly to find the reference.
This reference structure is found through the highest combined similarity between an incomplete and complete assembly.
The structural feature similarity between two structures \( \mathbf{V}_1 \) and \( \mathbf{V}_2 \) is defined as:
\begin{equation}\label{eq:similarity}
\text{S}_{\text{feat}}(\mathbf{V}_1, \mathbf{V}_2) = \frac{1}{2} \left( \text{S}_{\text{comp}} + \text{S}_{\text{dens}} \right)
\end{equation}
where
\begin{equation}
\text{S}_{\text{comp}} = 1 - \frac{| \text{CC}(\mathbf{V}_1) - \text{CC}(\mathbf{V}_2) |}{\max(\text{CC}(\mathbf{V}_1), \text{CC}(\mathbf{V}_2))} \notag
\end{equation}
\begin{equation}
\text{S}_{\text{dens}} = 1 - \frac{| \rho(\mathbf{V}_1) - \rho(\mathbf{V}_2) |}{\max(\rho(\mathbf{V}_1), \rho(\mathbf{V}_2))} \notag
\end{equation}

Here, \(\text{CC}(\mathbf{V})\) is the number of connected components in the structure \( \mathbf{V} \), and \( \rho(\mathbf{V}) \) is the voxel density and the combined similarity is defined as:
\begin{equation}\label{eq:combined_similarity}
\text{S}_{\text{com}}(\mathbf{V}_1, \mathbf{V}_2) = \alpha \cdot \exp(-\text{D}_{\text{CD}}(\mathbf{V}_1, \mathbf{V}_2)) + \beta \cdot \text{S}_{\text{feat}}(\mathbf{V}_1, \mathbf{V}_2)
\end{equation}
where \( \alpha \) and \( \beta \) are the weights for geometric and structural feature similarities, respectively, and \( \text{D}_{\text{CD}} \) is the Chamfer distance between \( \mathbf{V}_1 \) and \( \mathbf{V}_2 \). 
The Chamfer distance (\(\text{D}_{\text{CD}}\)) measures geometric similarity through the average distance between points on the surfaces of two voxels. 

\vspace{-1pt}
\subsection{Combinatorial Deep Reinforcement Learning}
\label{sec:methodDRL}
This section introduces our proposed framework of combinatorial DRL for the completion task.
We explain how our agent learns from an incomplete and complete structure to select optimal actions at each sequential step.
In addition, we propose a validity action mask to filter invalid actions and demonstrate it in a discrete Lego assembly environment. Refer to the right side of \cref{fig:overall_pipeline} for the DRL pipeline.



\textbf{State:}
For every $t$-th state $s_t$, the MDP is defined as $s_t=(\mathcal{V}^{t}_{cur}, \mathcal{V}_{\text{tar}}, \mathcal{I}_t)$ as represented at the top right of \cref{fig:overall_pipeline}. 
$\mathcal{V}^{t}_{cur}$ is a 3D binary matrix that represents the incomplete voxel being assembled. $\mathcal{V}_{tar}$ is a 3D binary matrix voxel that represents the complete target voxel that $\mathcal{V}^{t}_{cur}$ is trying to replicate.
$\mathcal{I}_t$ is a dictionary that tracks the assembly inventory of available bricks.

\textbf{Action:} At each $t$-th state $s_t$, we define an action space of all feasible placements of Lego bricks within the current voxel matrix $\mathcal{V}^{t}_{cur}$.
Since $\mathcal{V}_{tar}$ is not a predefined assembly blueprint but rather a reference for constructing missing features, the agent has to output a decision $a_t=(B_t, p_t, \omega_t)$ in a vast number of invalid and valid actions generated by the action space, which the actor $\pi$ determines at a given $t$-th step in order to satisfy sequential assembly:
$B_t\in [1, N]$ defines the brick type/size. $p_t=(x_t, y_t, z_t), x_t\in[1,h], y_t\in[1,w], z_t\in[1,d],$ defines the brick's position within the action space.
$\omega_t\in[0, 1]$ defines either a horizontal or vertical orientation for each Lego brick.
In choosing a valid action, we define a validity action mask, which utilizes heuristics to mask out all invalid action probabilities from the probability distribution that the actor selects from.

\textbf{Transition Function:}
The state transition function $\mathcal{P}$ maps the current state $s_t$ and action $a_t$ to the next state $s_{t+1}$, updating the current voxel representation $\mathcal{V}^{t}_{\text{cur}}$ with the newly placed brick and adjusting the brick inventory accordingly. 
The brick inventory is represented as $\mathcal{I}_{t+1}^{B_t}=\mathcal{I}_{t}^{B_t}-1$.

Formally, this can be represented as:
$s_{t+1} = \mathcal{P}(s_t, a_t) = (\mathcal{V}^{t}_{\text{cur}} \cup a_t, \mathcal{V}_{\text{tar}}, \mathcal{I}_{t+1}).$
The state transition accounts for the addition of the selected brick in the specified position and orientation and the decrement in its inventory count.

\textbf{Reward Function:}
The goal for 3D Assembly Completion is to construct an object $\mathcal{V}_t$ that is as similar as possible to the target object $\mathcal{V}_{tar}$.
Thus, we define the accumulated reward as a measure of the space utilization of the target voxel as well as the similarity metric defined in \cref{eq:combined_similarity}:

\begin{equation}\label{eq:reward}
 r(a_t, s_t) =\mathcal{V}^t_{cur}\cap\mathcal{V}_{tar}\times{\textit{c}}+S_{com}\times{\textit{d}}
\end{equation}
where the first half of the equation represents the number of overlapping voxels between the current and target voxel grid. c and d represent weight parameters that guide the agent’s learning.

\subsubsection{Action Validity Mask}
To quantify why training DRL for combinatorial 3D assembly completion is difficult, we want to emphasize that we define our environment as a 48x48x48 width, height, and depth, along with 8 unique bricks and 2 orientations for each brick. This would mean that at each state $s_t$, the action space would total up to 48*48*48*8*2=1,769,472 unique actions. 
Similar works addressed at combinatorial planning like \cite{ChungH2021neurips, GraphTransformerASP} have an exact end voxel to fill out, whereas, in assembly completion, the end target is not exact, which creates a greater need for invalid action filtering and constraining heuristics. 
If ASP heuristics or only a reward is used, training the agent is impossible due to the sheer number of actions created from the assembly completion task. 
Therefore, we consider multiple heuristics that prevent invalid actions such as floating, unstable bricks, or colliding bricks.

\textbf{Object Boundary:}
To elaborate further,  the agent has no definitive done condition (\eg  $\mathcal{V}_{cur}$ == $\mathcal{V}_{tar}$) and needs the ability to place bricks outside of $\mathcal{V}_{tar}$ to achieve a higher similarity, whereas traditional ASP works hard-constrain the action space to $\mathcal{V}_{tar}$. 
Therefore, by scaling and aligning in \cref{infer&match}, we assume the target voxel $\mathcal{V}_{tar}$ to roughly match the incomplete assembly and define a tolerance boundary so that bricks further outside the target area are not considered.
This is defined as:
\begin{equation}
 O(a_t, s_t, s_{t+1})=
 \mathcal{V}_{\text{tar}} \cup \text{tol}(\mathcal{V}_{\text{tar}}, X) \}
\end{equation}
The function $\text{tol}(\mathcal{V}_{\text{tar}}, X)$ generates a boundary of $X$ grid spaces outward from the target voxel around the entire object itself.

\textbf{Collision:}
Since real-world assembly does not allow for collisions between objects, we filter out all possible actions that intersect with any existing bricks.

\textbf{Inventory:}
Since real-world assembly has a finite inventory, the agent may only use a certain brick type if still available.
\begin{equation}
    \begin{split} 
    I(a_t, s_t, s_{t+1})= \mathcal{I}_{t}^{B_t}>0,
    \end{split}
\end{equation}

\textbf{Hooking Mechanism:}
As illustrated in \cref{fig:constraints}, a brick $B_t$ is only considered valid if there exists one adjacent brick $B_{t-1}$ such that the bottom face of $B_t$ is directly above or the top face of $B_t$ is hanging below.

Mathematically, the hooking condition can be expressed as: 
\begin{equation}
 H(a_t, s_t) = 
 \begin{cases}
(x_t\ldots x_t+h, y_t\ldots y_t+w, z_t-1) \in B_{t-1}
\\
(x_t\ldots x_t+h, y_t\ldots y_t+w, z_t + 1) \in B_{t-1}
\end{cases}
\end{equation}

Additionally, the hooking mechanism restricts the placement of bricks to eliminate floating bricks which are not feasible in practical scenarios.
By enforcing this rule, we guide the agent to construct more practical 3D assemblies.

\textbf{Similarity Constraint:}
Even after multiple heuristic constraints, a vast number of actions remain unfiltered. 
Thus, we constrain that the placed brick can not worsen the similarity between $\mathcal{V}^{t}_{cur}$ and $\mathcal{V}_{tar}$.
The goal is to ensure that at each timestep \textit{t}, the bricks continuously attempt to replicate the structural features of  $\mathcal{V}_{tar}$ and guarantee progress:
\begin{equation}
    S(a_t, s_t, s_{t+1}) = S_{com}( \mathcal{V}^{t}_{cur}, \mathcal{V}_{tar}) + \beta < S_{com}( \mathcal{V}^{t+1}_{cur}, \mathcal{V}_{tar})
\end{equation}

where $\mathcal{S}_{com}$ is defined in \cref{eq:combined_similarity} and $\beta$ is a tolerance threshold.

\section{Experiments}
\begin{figure} [H]
\centering
\subfigure[]{\includegraphics[width=0.17\linewidth]{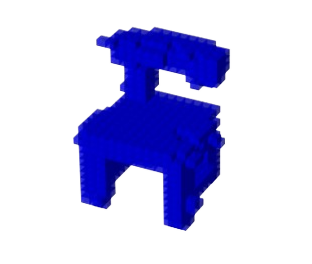}\label{fig:h1}}\hfill
\subfigure[]{\includegraphics[width=0.17\linewidth]{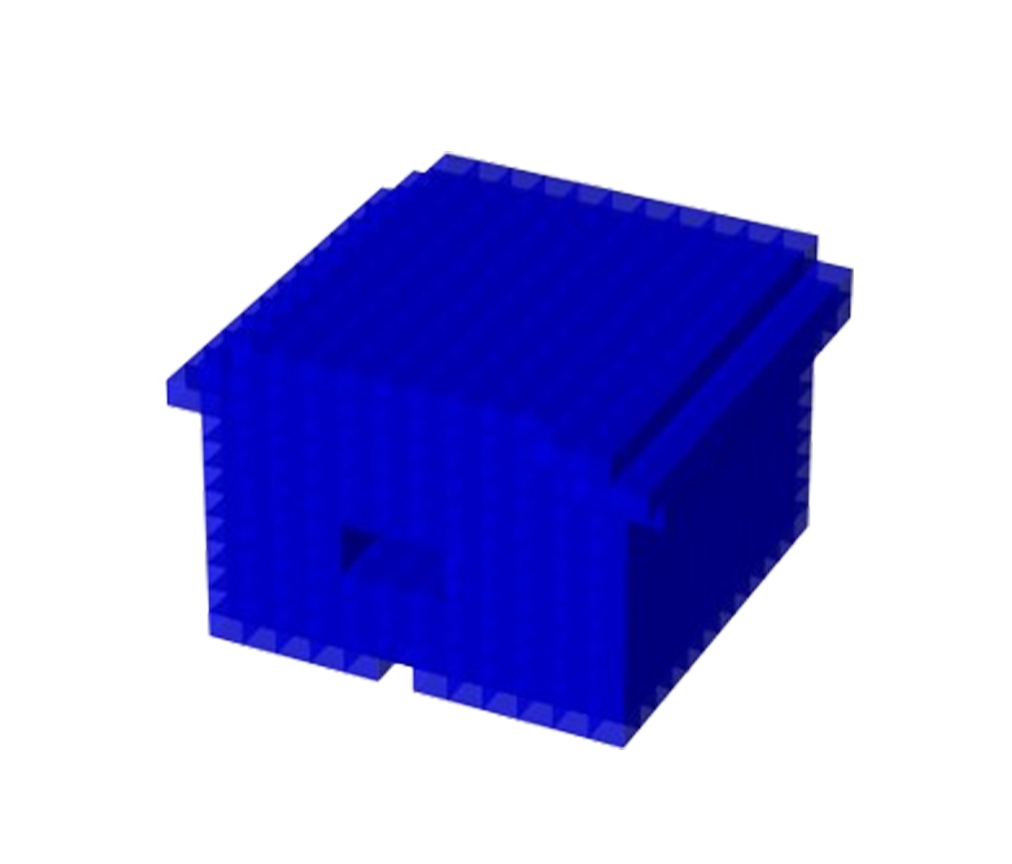}\label{fig:h2}}\hfill
\subfigure[]{\includegraphics[width=0.17\linewidth]{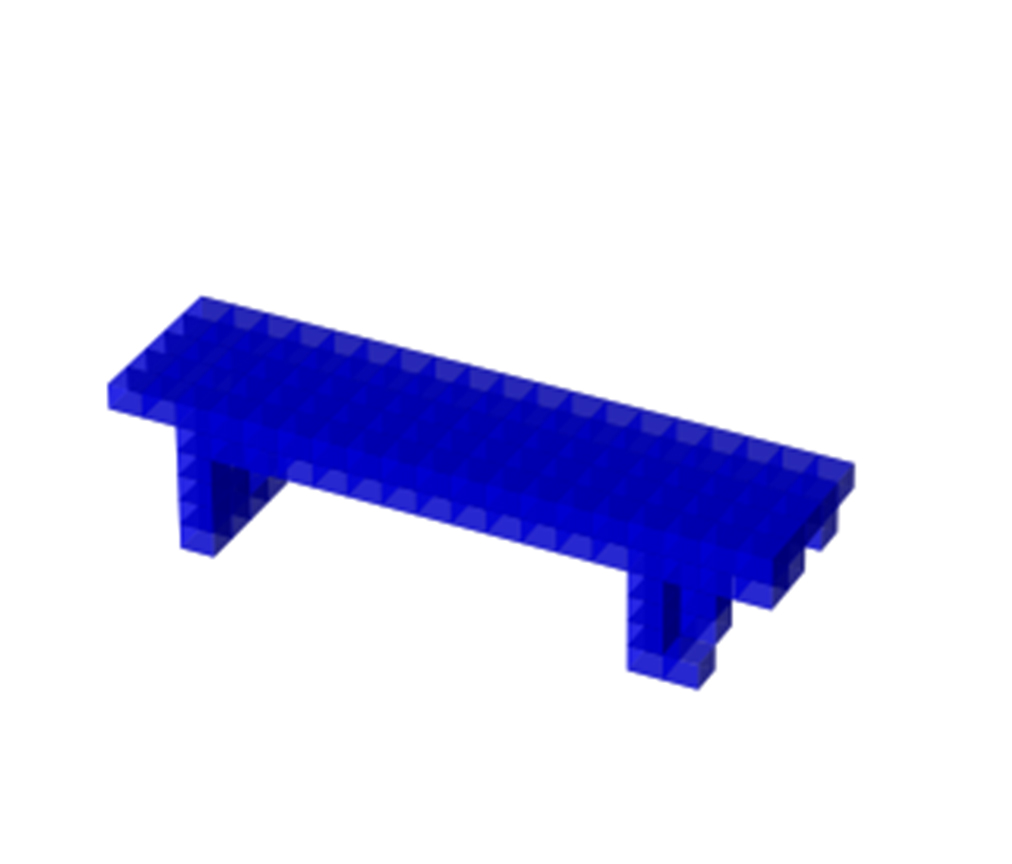}\label{fig:h3}}\hfill
\subfigure[]{\includegraphics[width=0.17\linewidth]{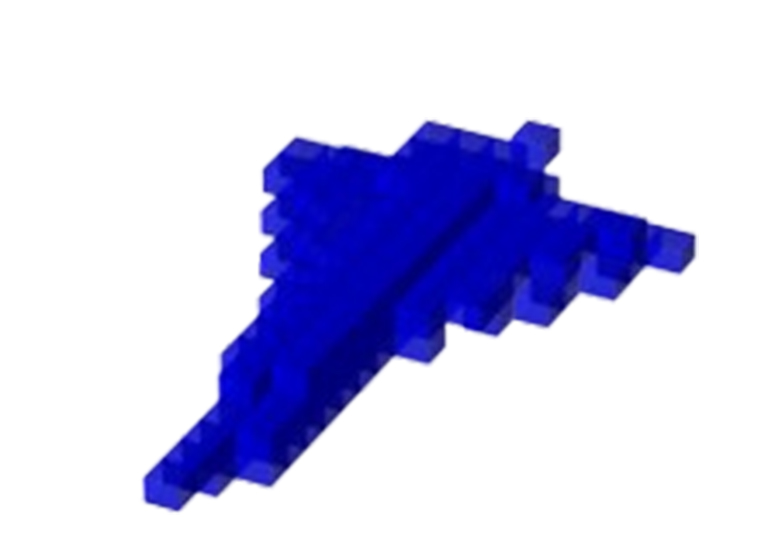}\label{fig:h4}}\hfill
\subfigure[]{\includegraphics[width=0.17\linewidth]{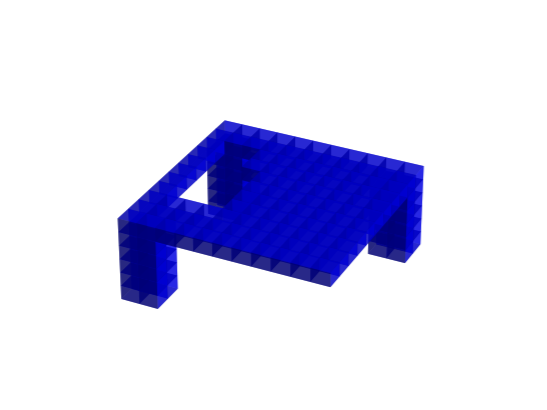}\label{fig:h5}}\hfill
\\
\vspace{-10pt}
\subfigure[]{\includegraphics[width=0.17\linewidth]{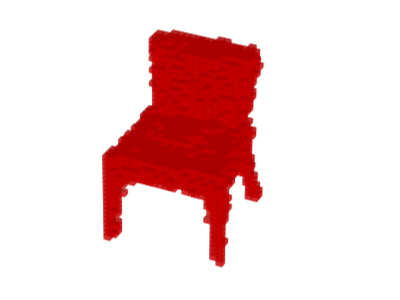}\label{fig:h6}}\hfill
\subfigure[]{\includegraphics[width=0.17\linewidth]{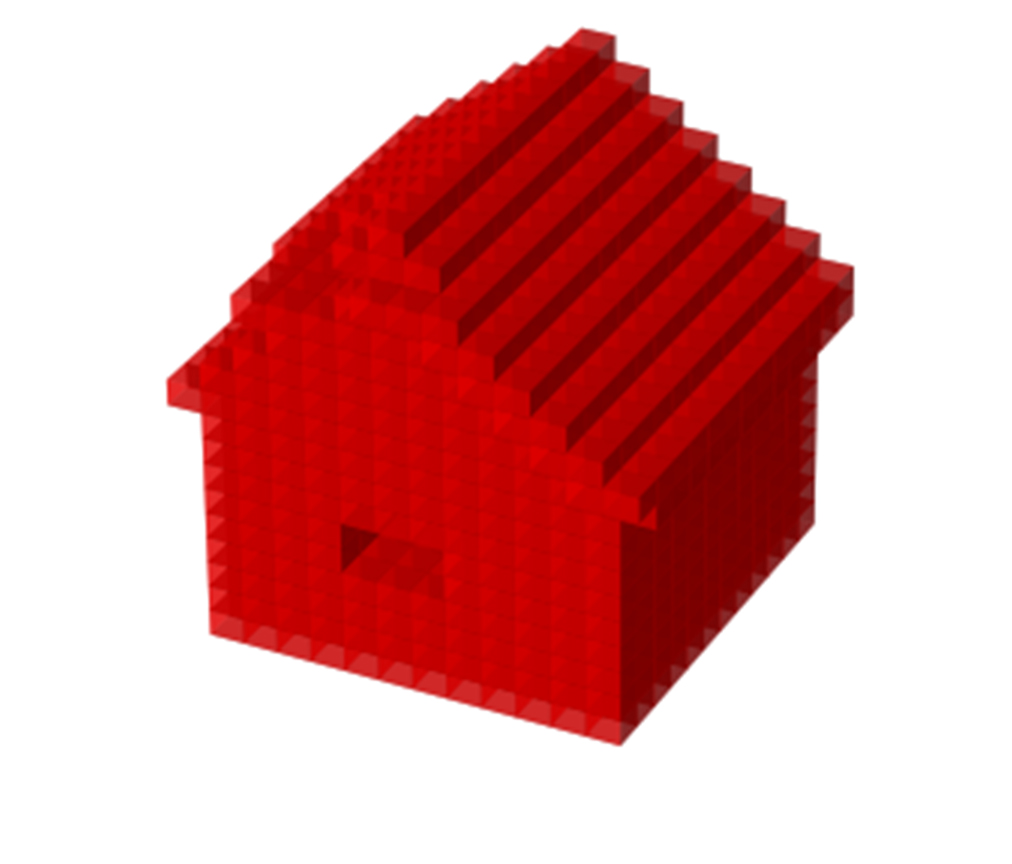}\label{fig:h7}}\hfill
\subfigure[]{\includegraphics[width=0.17\linewidth]{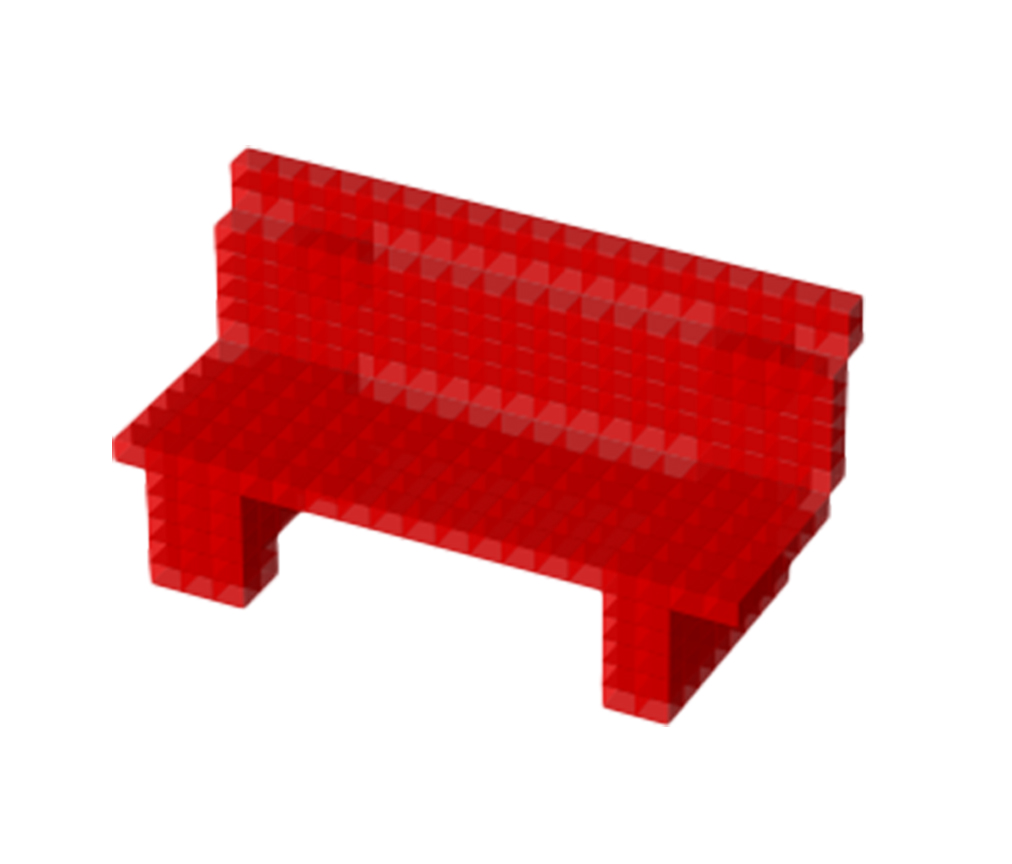}\label{fig:h8}}\hfill
\subfigure[]{\includegraphics[width=0.17\linewidth]{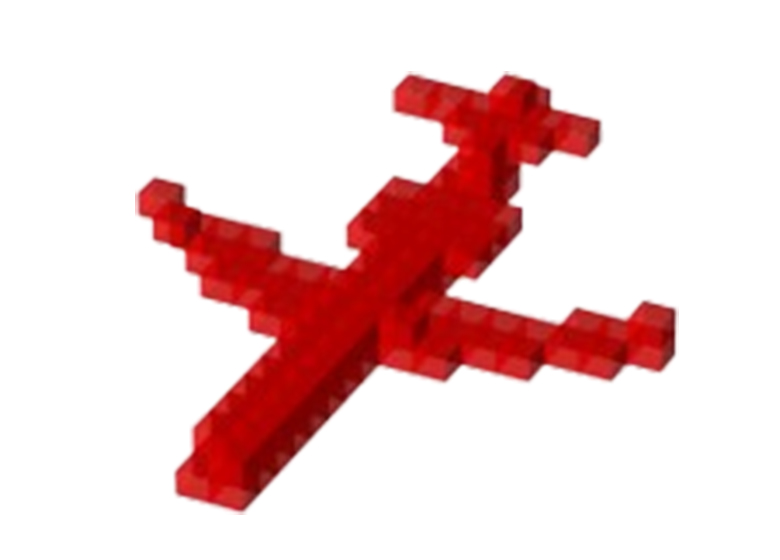}\label{fig:h9}}\hfill
\subfigure[]{\includegraphics[width=0.17\linewidth]{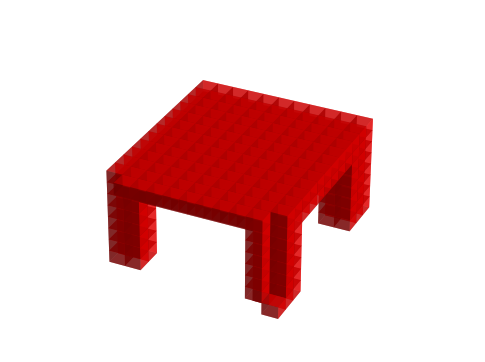}\label{fig:h10}}\hfill
\\
\vspace{-10pt}
\subfigure[]{\includegraphics[width=0.17\linewidth]{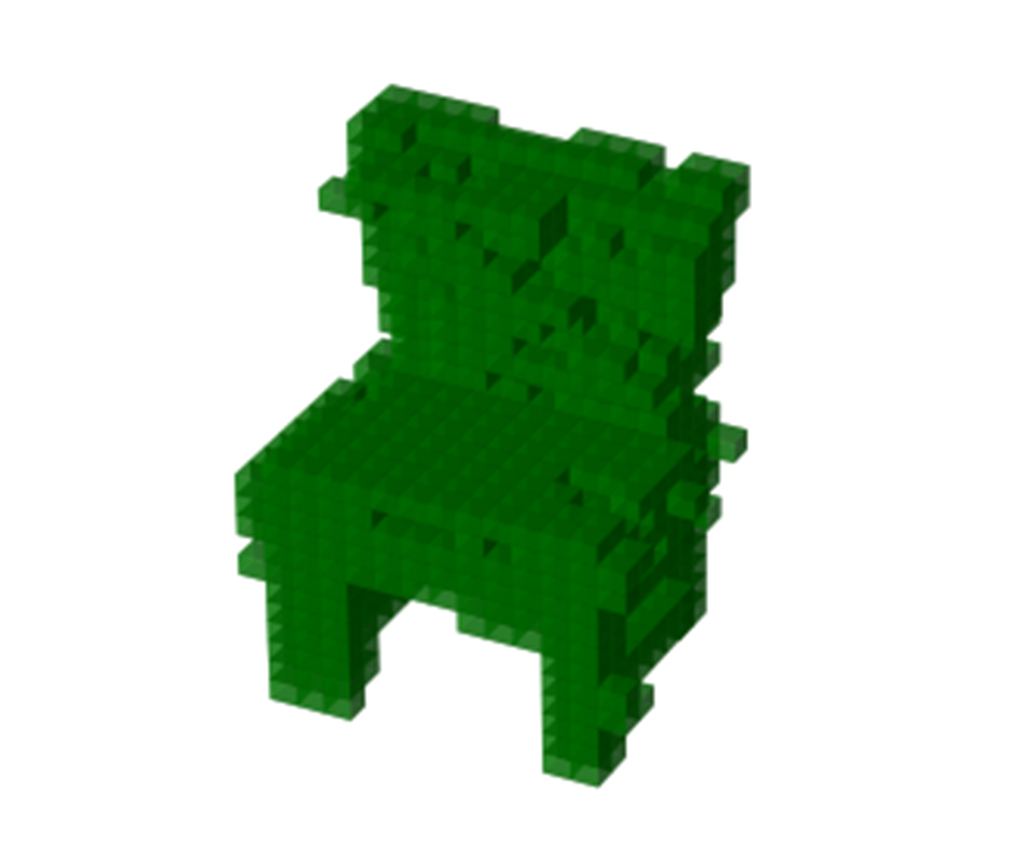}\label{fig:h11}}\hfill
\subfigure[]{\includegraphics[width=0.17\linewidth]{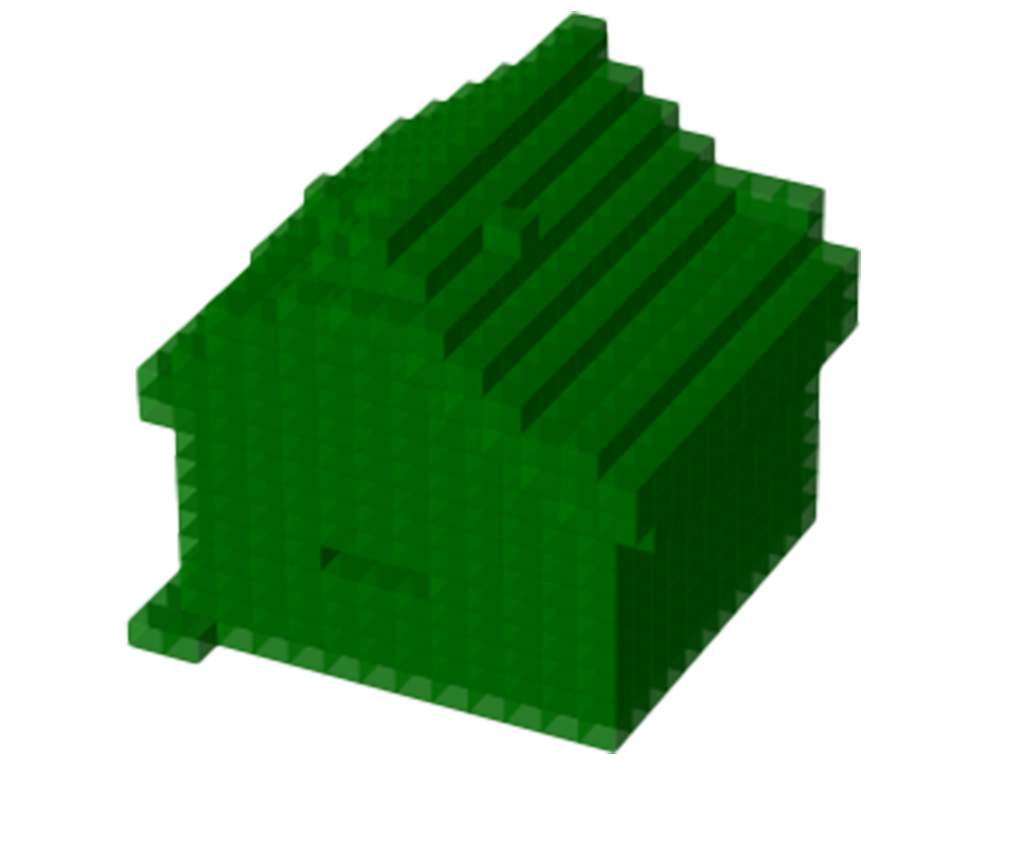}\label{fig:h12}}\hfill
\subfigure[]{\includegraphics[width=0.17\linewidth]{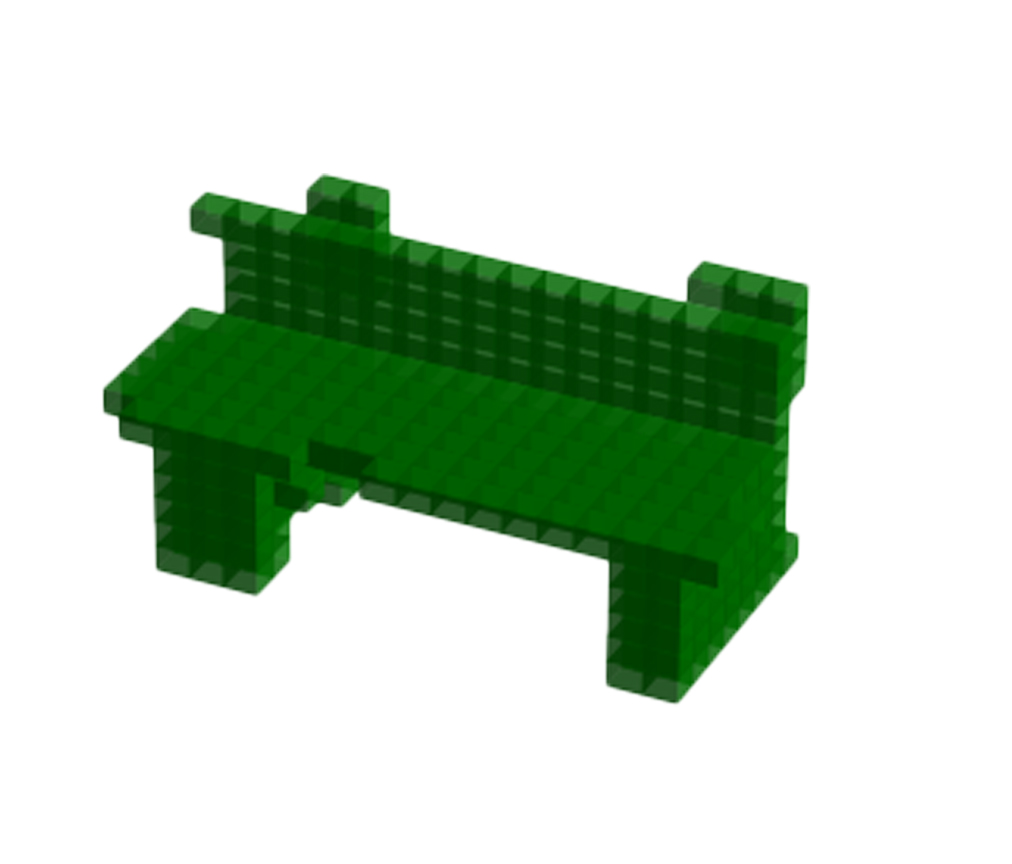}\label{fig:h13}}\hfill
\subfigure[]{\includegraphics[width=0.17\linewidth]{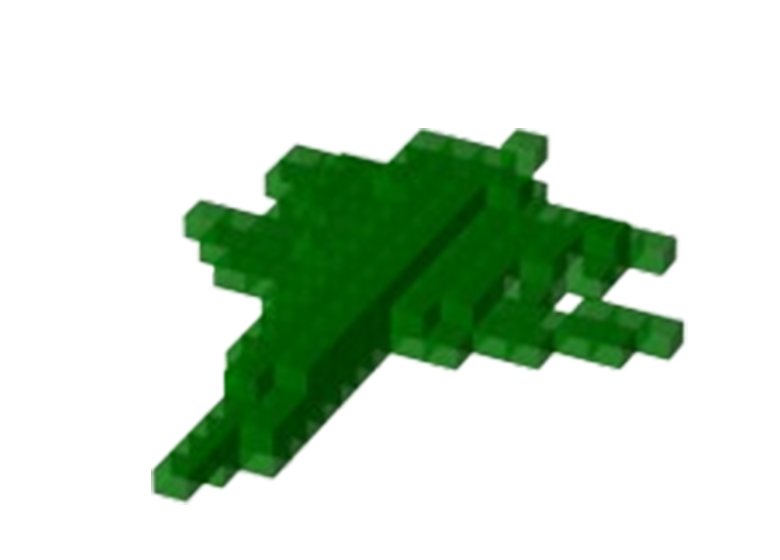}\label{fig:h14}}\hfill
\subfigure[]{\includegraphics[width=0.17\linewidth]{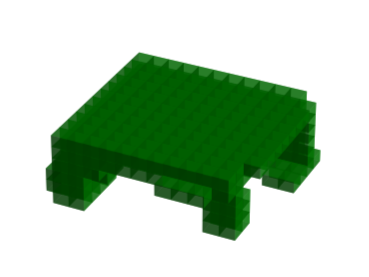}\label{fig:h15}}\hfill
\\
\vspace{-5pt}
    \caption{\footnotesize Top row: Incomplete structure input.
    Middle: Stable matched reference object; not to original scale/rotation. Bottom: Visualized Result. 
    Only \cref{fig:h1} has been seen before in training; the agent successfully completes other unseen incomplete structures.  
    \label{fig:exp1_5obj}}
    \vspace{-5pt}
\end{figure}

We evaluate the performance of our method through similarity and stability metrics defined in \cref{eq:combined_similarity}, \cite{liu2024stablelegostabilityanalysisblock} and visualize results in \cref{fig:exp1_5obj}.
Then, we perform two ablation studies on our proposed method to highlight the effectiveness of the robot inference and completion steps.
Finally, we evaluate our finished assemblies using stability analysis to demonstrate real-world feasibility and validate our approach.

To demonstrate our proposed framework, we define a 48x48x48 LEGO base plate environment with a brick storage inventory of 8 unique bricks (1x1, 1x2, 1x4, 1x6, 1x8, 2x2,  2x4, 2x6) and 2 orientations for each brick.
We evaluate our framework across multiple object categories and visualize the result. 
We implement PPO \cite{schulman2017proximalpolicyoptimizationalgorithms} for stable training and develop the action validity mask using \cite{TANG2020200} in Python.
All models are trained on a Ubuntu 20.04 PC with a 48-core Intel(R) Xeon(R) Silver 4214 CPU @ 2.2.GHz with an Nvidia RTX A4000 GPU with 16GB memory.

During training, the DRL agent takes a set of incomplete structures (\eg \cref{fig:h1}), finds the reference assembly with the highest similarity from a library (\eg \cref{fig:h6}), and then sequentially builds to complete the structure (\eg \cref{fig:h11}). 
Then, we test the model's performance on various unseen incomplete structures in multiple different object categories with dramatically different dimensions, shapes, complexities, etc. 

\vspace{-5pt}

\subsection{Qualitative Analysis}
The trained model demonstrates great generalizability during testing. 
In \cref{fig:exp1_5obj}, multiple unseen incomplete structures such as \cref{fig:h2} partial house, \cref{fig:h3} partial bench, \cref{fig:h4} partial plane and \cref{fig:h5} partial table are all dramatically different from the \cref{fig:h1} partial chair, which the agent was trained on. 
Regardless, the agent generalizes across unseen, unknown object categories and eliminates the need for model re-training. 
 
The model is scalable to different dimensions and overcomes rotational and scale invariance.
For example, the model can handle large assemblies like the incomplete structure \cref{fig:h1}, which is 14x12x20 and 287 bricks, while the matched chair in \cref{fig:h6} is 22x20x30 and 862 bricks.
For smaller assemblies, the incomplete structure in \cref{fig:h4} is 12x18x5 and 56 bricks, while the matched plane \cref{fig:h9} is 15x15x5 and takes 41 bricks to assemble. 

Moreover, the model demonstrates precise capabilities in inferring missing features from the reference structure to sequentially build the missing features onto the incomplete structure. 
In \cref{fig:h2}, \cref{fig:h7}, \cref{fig:h12}, the agent infers from a house with an unfinished roof and smaller base, then completes the roof feature using a reference of the matched structure \cref{fig:h7}.
In \cref{fig:h4}, \cref{fig:h9}, \cref{fig:h14}, the agent infers from an incomplete plane object, finds the best reference structure \cref{fig:h9} from the object library, and builds missing wings to create \cref{fig:h14}. 
However, due to subtle feature mismatches between the incomplete structure and the reference plane combined with stability constraints in the action space, the agent could not construct the missing tail feature as shown in \cref{fig:h9}. 
Further relaxation of the heuristic constraints and employing more advanced assembly techniques might allow the agent to complete the missing tail but would require longer processing time due to the expanded action space.  

\subsection{Quantitative Evaluation}
In this part, we evaluate and showcase the structure's similarity \cref{eq:combined_similarity} and stability.

\begin{table} [H]
\centering
\begin{tabular}{c c  c c  } 
\hline
LEGO Object & Initial Similarity  \cref{eq:combined_similarity} & Final Similarity & Stable \\ 
\hline
Plane   &  67.7\% & 73.2\% & $\checkmark$\\
House  & 74.7\% &  77.5\% & $\checkmark$\\
Bench    & 82.8\% &  94.9\% & $\checkmark$\\
Table    &  82.5\% & 98.8\% & $\checkmark$\\
Chair    &  77.5\%  & 89.6\% & \ding{55}\\
\hline
\end{tabular}
\caption{\footnotesize Metric evaluation of multiple LEGO structures from \cref{fig:exp1_5obj} \label{table:comparison_dv}: The agent can complete all these structures and improve their similarity to their corresponding reference structures.}
\end{table}
\vspace*{-\baselineskip}

The initial and final similarity percentages indicate how closely the reconstructed structure matches the reference in \cref{table:comparison_dv}.




For example, the chair assembly starts at 77.5\% and improves to 89.6\% as it handles moderate complexity in structures and completes the missing backrest feature.

The incomplete table assembly shows the most significant improvement from 82.5\% to 98.8\% as the agent completes a key missing leg feature and tabletop surface part. 
This indicates that the framework excels in reconstructing simpler structures with fewer missing features, achieving near-complete similarity with the reference.

\subsection{Ablation Study}

\setlength{\intextsep}{2pt}     

\begin{table} [H]
    \centering
    \begin{tabular}{|c|c|c|c|c|}
        \hline
        \cline{2-5}
        &  No Scale & RANSAC Only & ICP Only
        & RANSAC\&ICP\\
        \hline
        Chair  & \ding{55} & 45\% & \ding{55} & 100\%\\
        \hline
        House  & \ding{55} & 100\% & \ding{55} & 100\% \\
        \hline
        Bench  & \ding{55} & 55\% & \ding{55} & 100\% \\
        \hline
        Plane  & \ding{55} & 40\% & \ding{55} & 100\%  \\
        \hline
    \end{tabular}
    \caption{\footnotesize 
    Effects of different point cloud manipulation components across 20 trials. Scaling is included in methods (2-4).}
    \label{table:ablation1}
\end{table}

Table \ref{table:ablation1} highlights the impact of different point cloud manipulation strategies on object alignment across 20 trials. 
Without scaling, no alignment is achieved. 
With only RANSAC, alignment success is inconsistent and suffers from larger incomplete and matched assembly differences. 
Only ICP fails to align any objects, emphasizing its dependence on prior rough alignment. 
However, combining RANSAC with ICP consistently achieves perfect alignment, initializing a perfect state for the DRL agent to operate on.

\vspace{10pt}
\begin{table} [H]
    \centering
    \begin{tabular}{|c|c|c|c|c|}
        \hline
        \multirow{2}{*}{} & \multicolumn{2}{c|}{Stability Mask} & \multicolumn{2}{c|}{No Stability Mask} \\
        \cline{2-5}
        & Similarity & Time (s)
        & Similarity & Time (s)\\
        \hline
        Tiny Chair (17) & 100\% & 158.4 & 100\% & 11.99 \\
        \hline
        Temple (54) & 95.9\% & 772.1 & 95.9\% & 20.92 \\
        \hline
        Bench (184) & \ding{55} & \ding{55} & 94.9\% & 353.38 \\
        \hline
        Big Chair (587) & \ding{55} & \ding{55} & 89.6\% & 3412.95 \\
        \hline
    \end{tabular}
    \caption{\footnotesize Comparison of time and similarity of different methods; (.) indicates the number of bricks the reference structure has; Time refers to total assembly duration, including robot inference step.}
    \label{table:ablation2}
\end{table}

Table \ref{table:ablation2} compares stability mask versus non-stability mask performance in terms of time and similarity.
We implement stability \cite{liu2024stablelegostabilityanalysisblock} into the stability mask similarly to \cite{ASPCoRL}.
Results show that excluding stability analysis significantly reduces computation time and has minimal effect on performance.
The stability mask approach is more feasible with smaller assemblies but fails with larger, more complex structures due to its computational demands. 
Conversely, the non-stability mask demonstrates good stability across most objects, though it does face stability challenges with very large assemblies, as seen with the 587 brick chair shown in \cref{table:comparison_dv} and \cref{fig:h11}.
Regardless, the non-stability mask can assemble a stable 184-brick bench that times out with a stability mask.
This demonstrates the effectiveness of the defined heuristics in \cref{sec:methodDRL} and the stable reference assembly, in which the agent integrates the stable reference features into the incomplete assembly.

\subsection{Stability Analysis}
\begin{figure}
\centering
\subfigure[]{\includegraphics[width=0.24\linewidth]{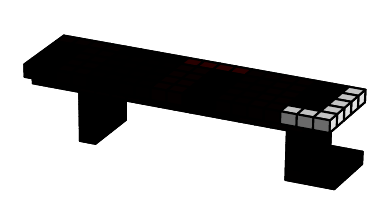}\label{fig:bench1}}\hfill
\subfigure[]{\includegraphics[width=0.24\linewidth]{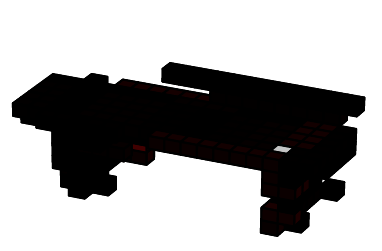}\label{fig:bench2}}\hfill
\subfigure[]{\includegraphics[width=0.24\linewidth]{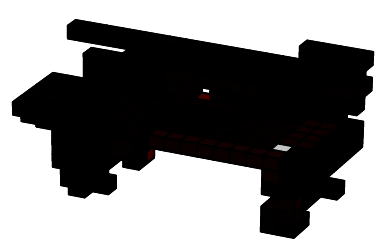}\label{fig:bench3}}
\subfigure[]{\includegraphics[width=0.24\linewidth]{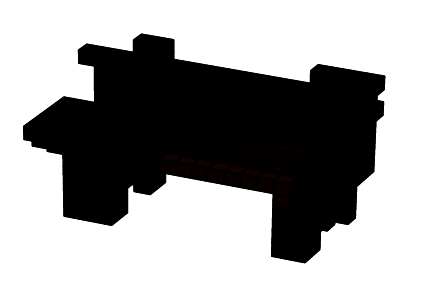}\label{fig:bench4}}\hfill
\\
\vspace{-10pt}
\subfigure[]{\includegraphics[width=0.24\linewidth]{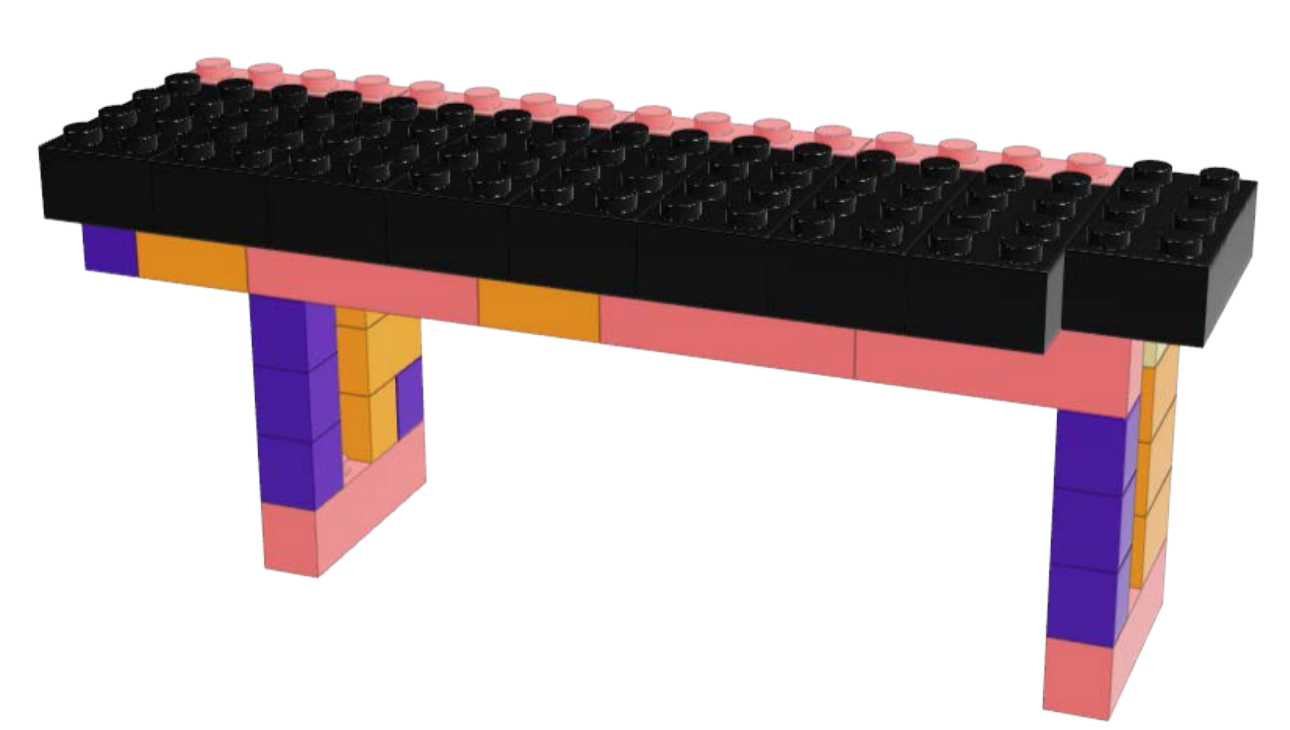}\label{fig:real_bench1}}\hfill
\subfigure[]{\includegraphics[width=0.24\linewidth]{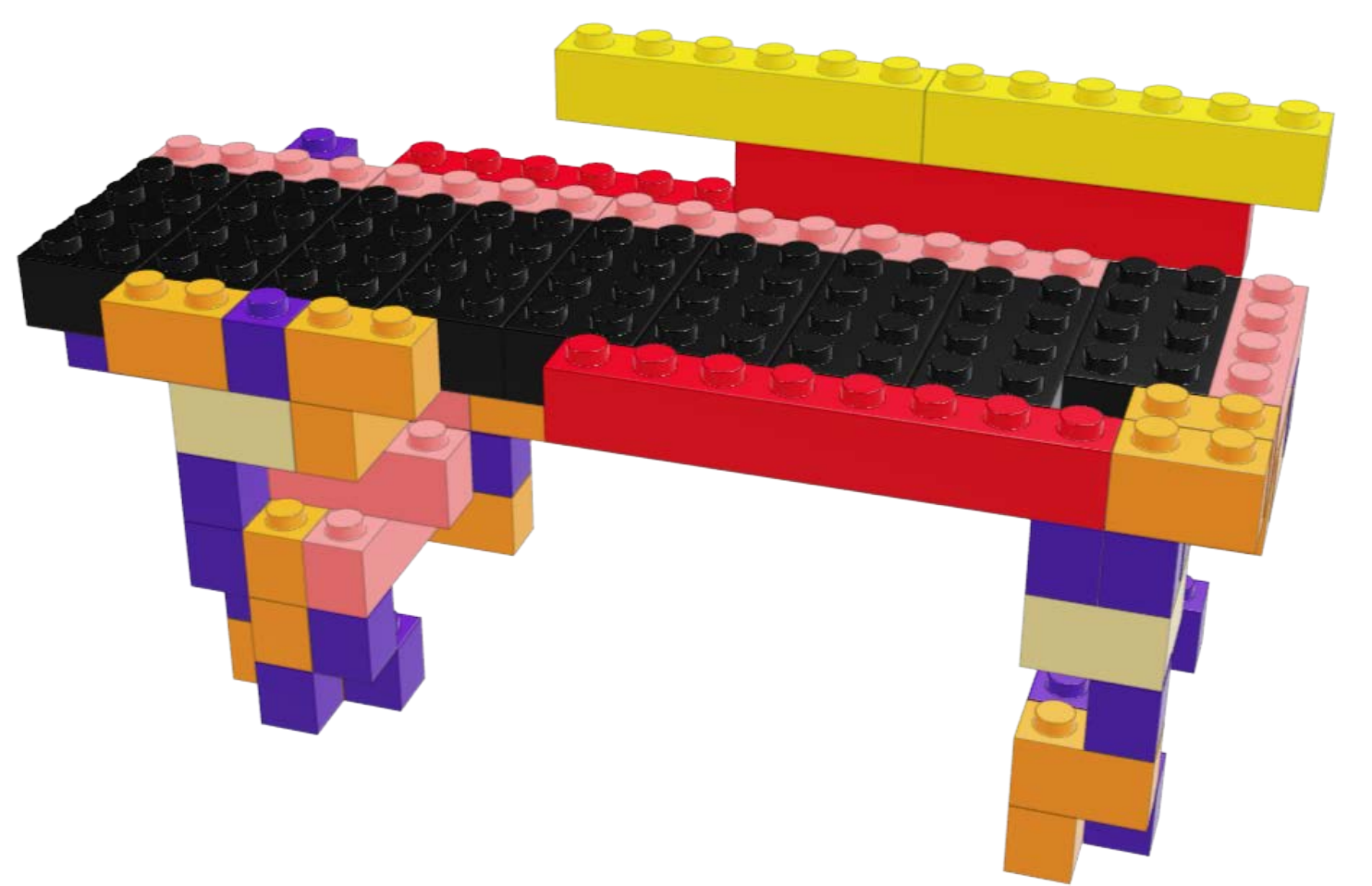}\label{fig:real_bench2}}\hfill\
\subfigure[]{\includegraphics[width=0.24\linewidth]{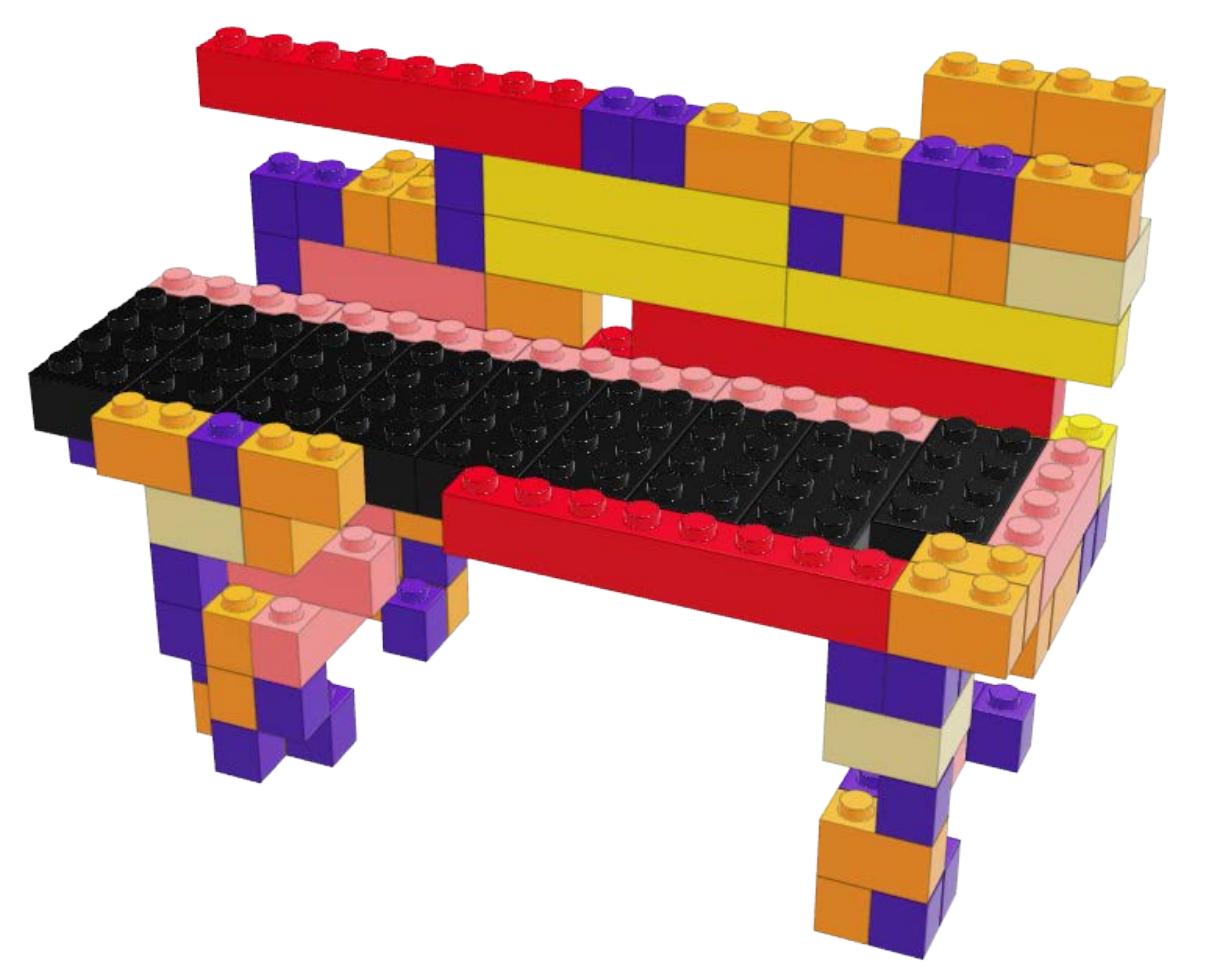}\label{fig:real_bench3}}\hfill
\subfigure[]{\includegraphics[width=0.24\linewidth]{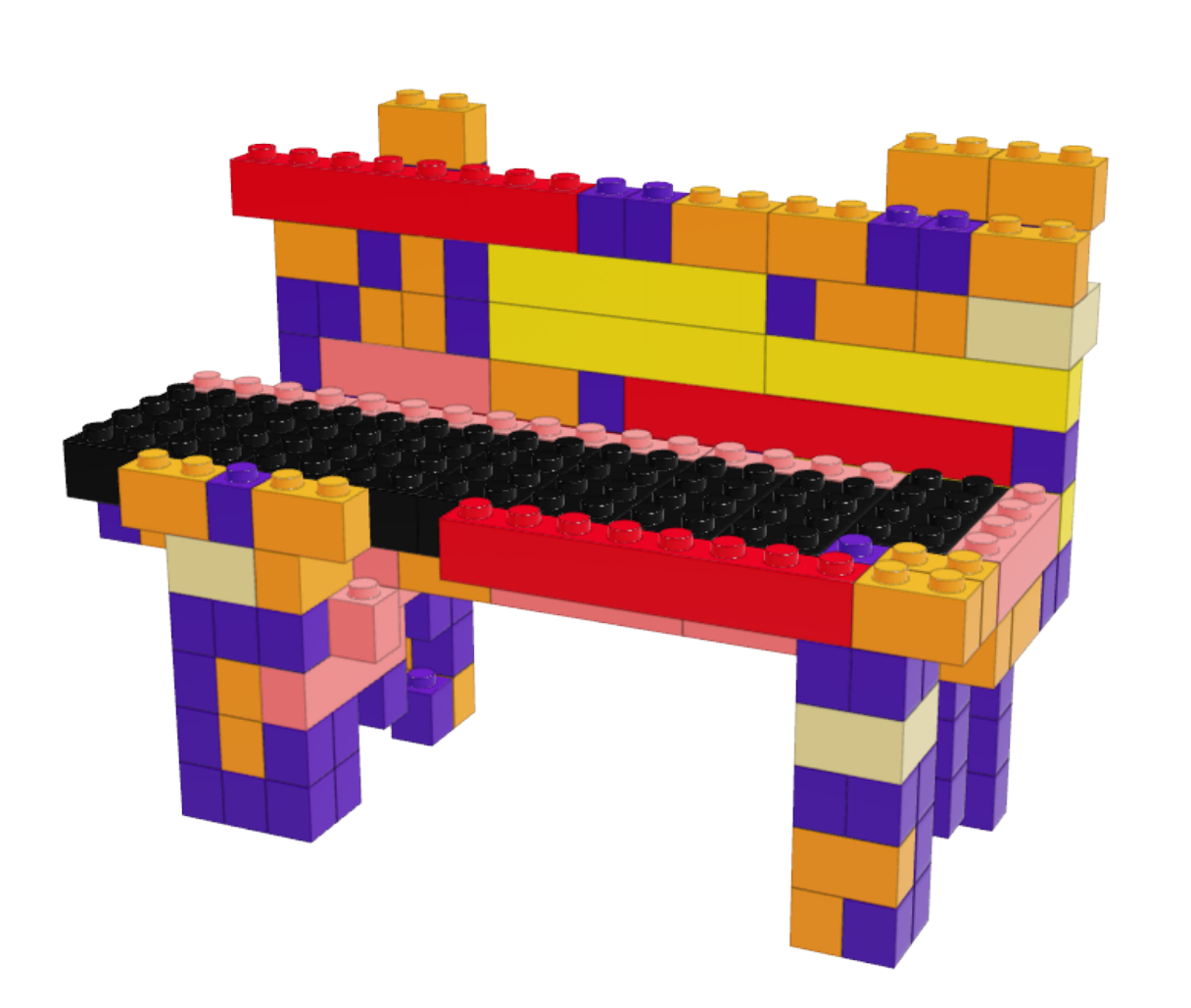}\label{fig:real_bench4}}\\
\vspace{-10pt}
    \caption{\footnotesize Example of sequential assembly with analysis. 
    Top: Stability Analysis, White-Collapsing, Red-Risky, Black-Stable. 
    Bottom: Real-world Lego Assembly, each unique color is a unique brick type.
    \label{fig:prototypes}}
    \vspace{-15pt}
\end{figure}

In this part, we evaluate the feasibility of our method through stability analysis \cite{liu2024stablelegostabilityanalysisblock} to assess the structural integrity throughout the construction sequences.
In \cref{fig:prototypes}, the incomplete bench assembly initially has unstable, collapsing bricks which are reflected in the bottom assembly.
However, after being matched with a stable reference bench and the assembly progresses, the collapsing bricks are removed entirely, while red stress areas diminish in intensity as the agent adds leg support features to hold the new backrest being assembled. 
Our method effectively manages and compensates for high pressure and tension, demonstrating its capability to handle real-world assembly challenges.

\subsection{Multiple Missing Features}
\vspace{-10pt}
\begin{figure} [H]
\centering
\subfigure[1 Feature.]{\includegraphics[width=0.24\linewidth]{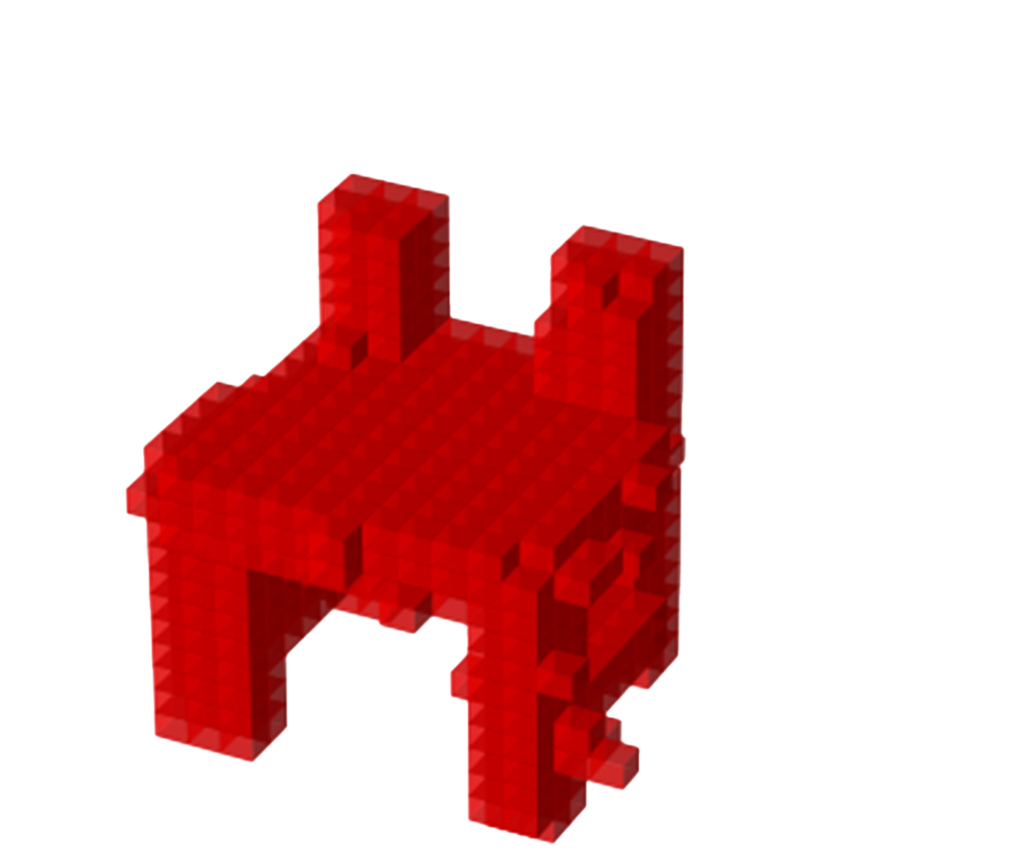}\label{fig:feat1}}
\subfigure[2 Features]{\includegraphics[width=0.24\linewidth]{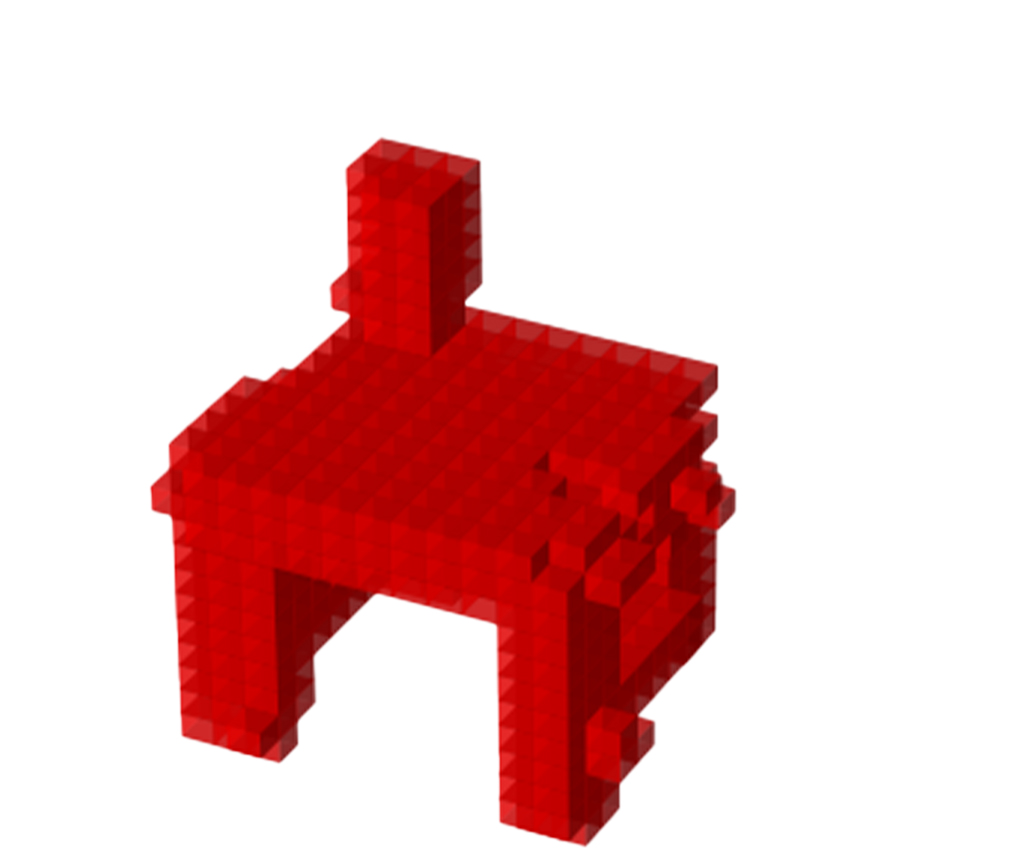}\label{fig:feat2}}\hfill
\subfigure[3 Features.]{\includegraphics[width=0.24\linewidth]{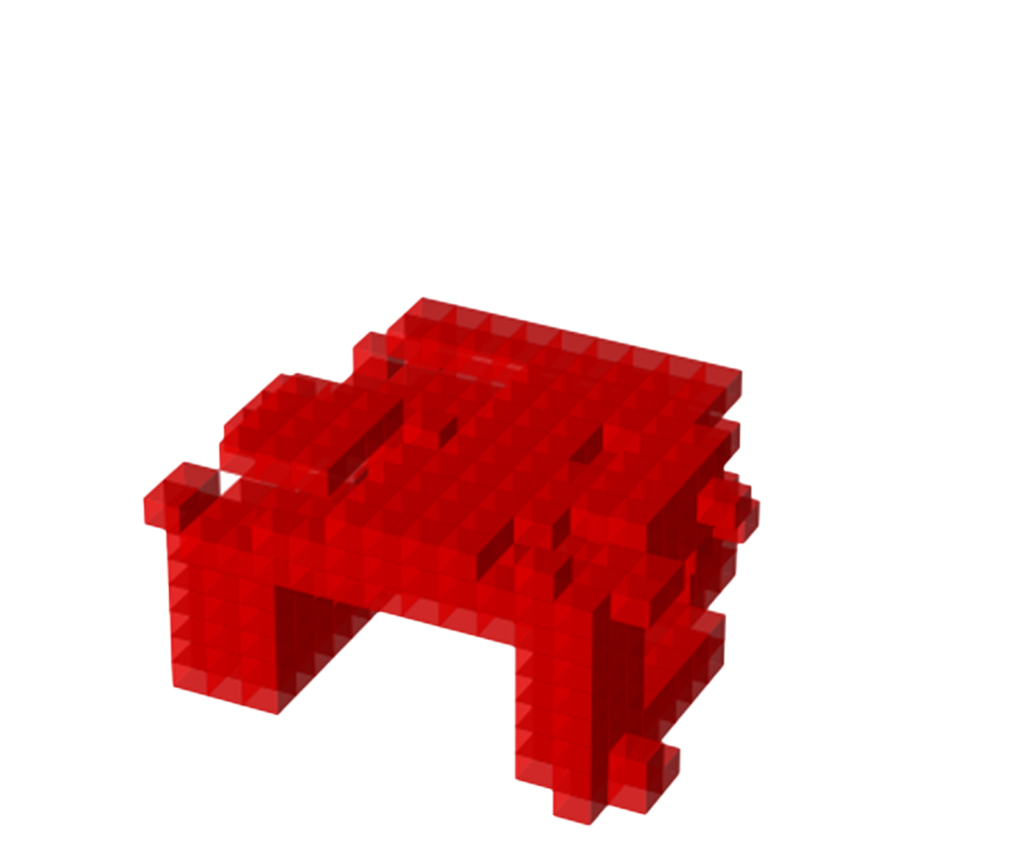}\label{fig:feat3}}\hfill
\subfigure[4 Features.]{\includegraphics[width=0.24\linewidth]{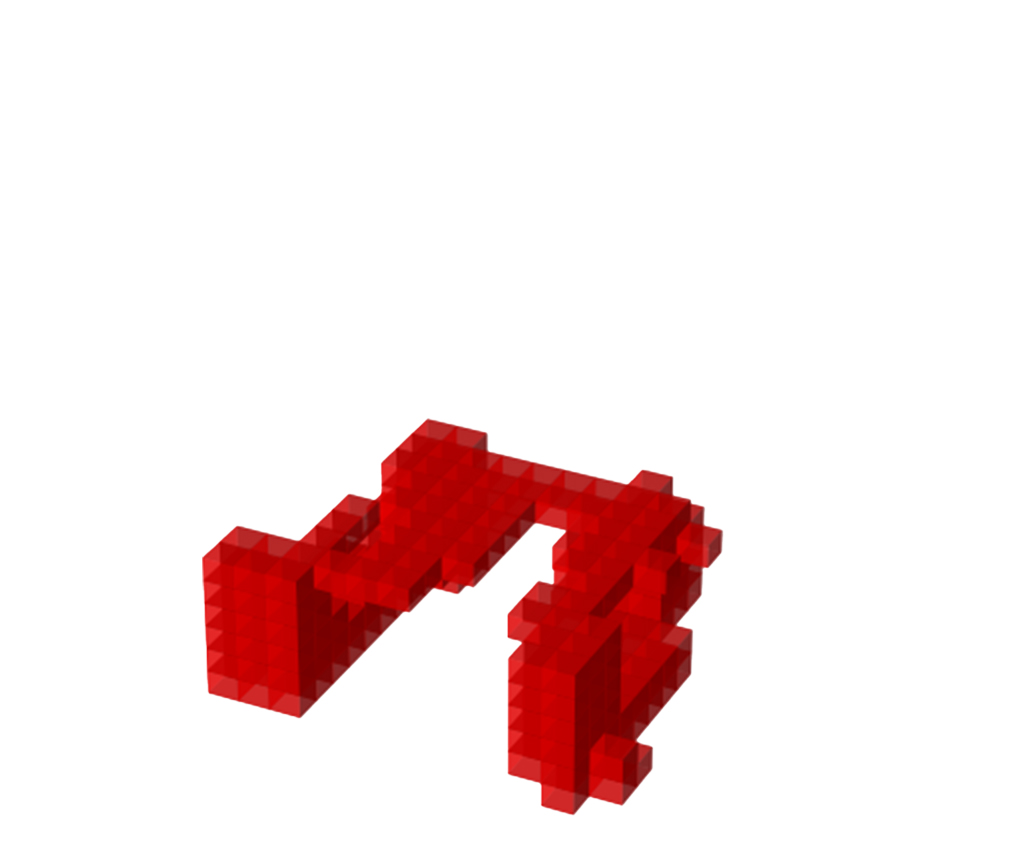}\label{fig:feat4}}\hfill
\\
\vspace{-5pt}
\subfigure[Chair 1.]{\includegraphics[width=0.24\linewidth]{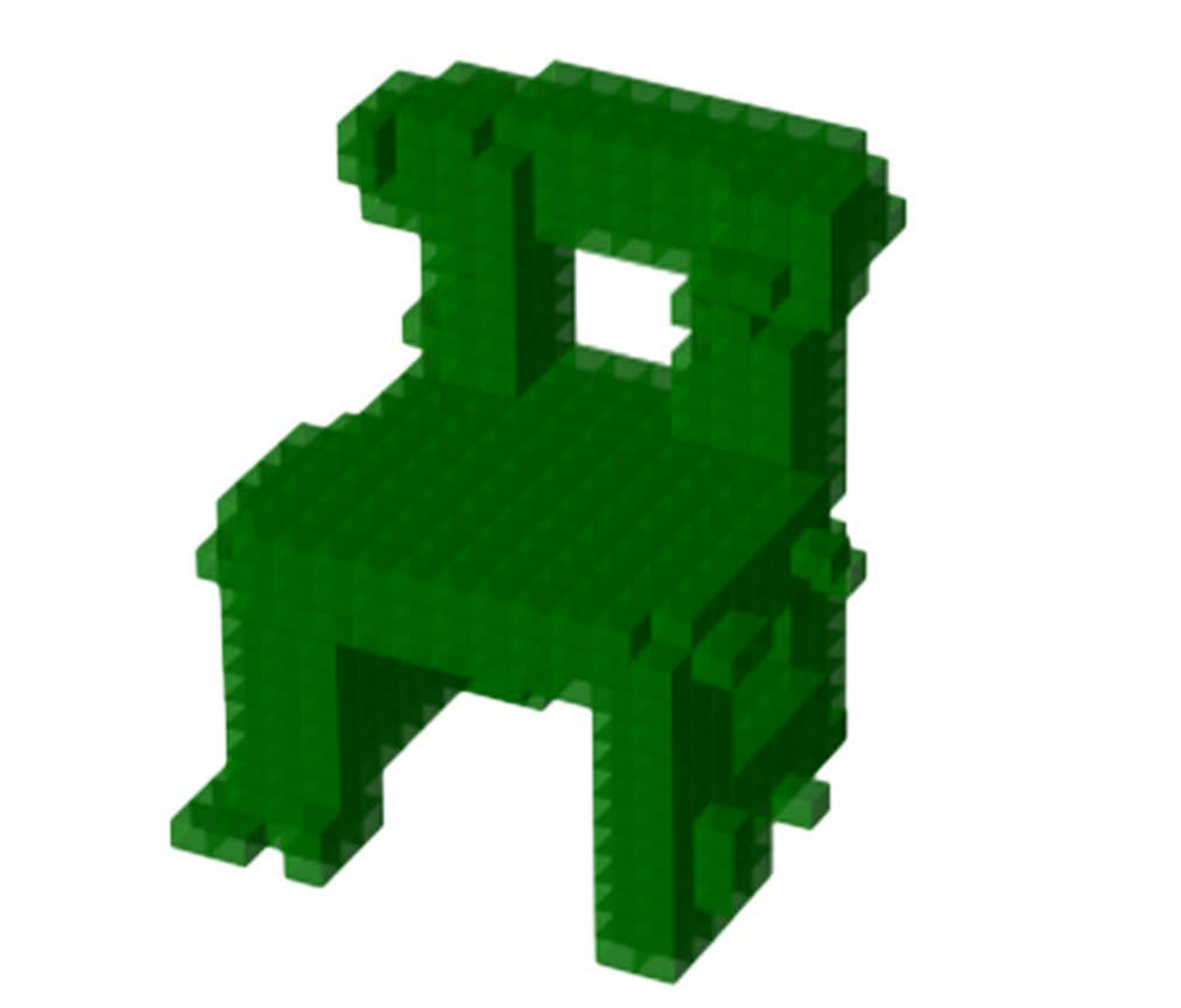}\label{fig:chair1}}
\subfigure[Chair 2.]{\includegraphics[width=0.24\linewidth]{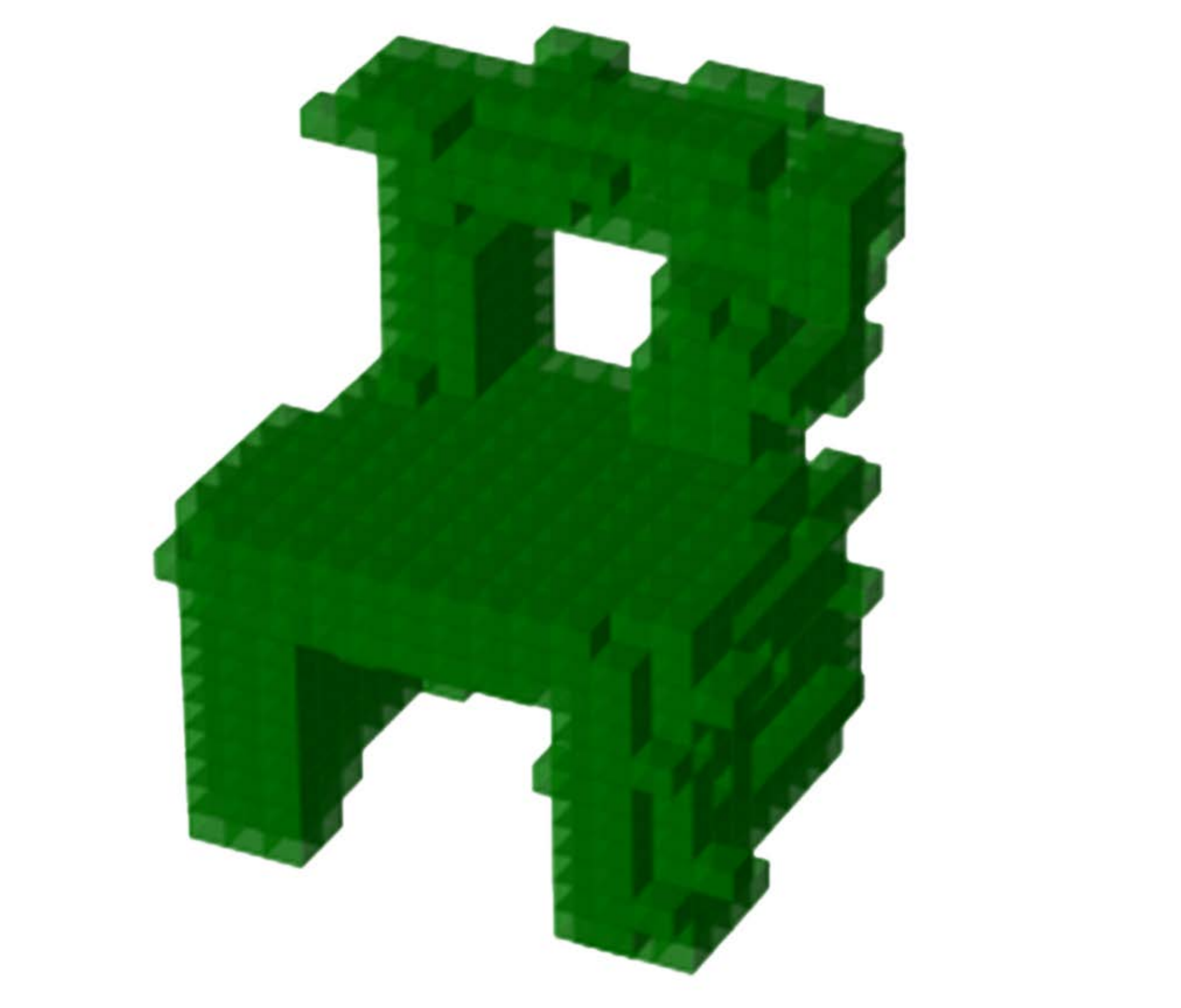}\label{fig:chair2}}\hfill
\subfigure[Chair 3.]{\includegraphics[width=0.24\linewidth]{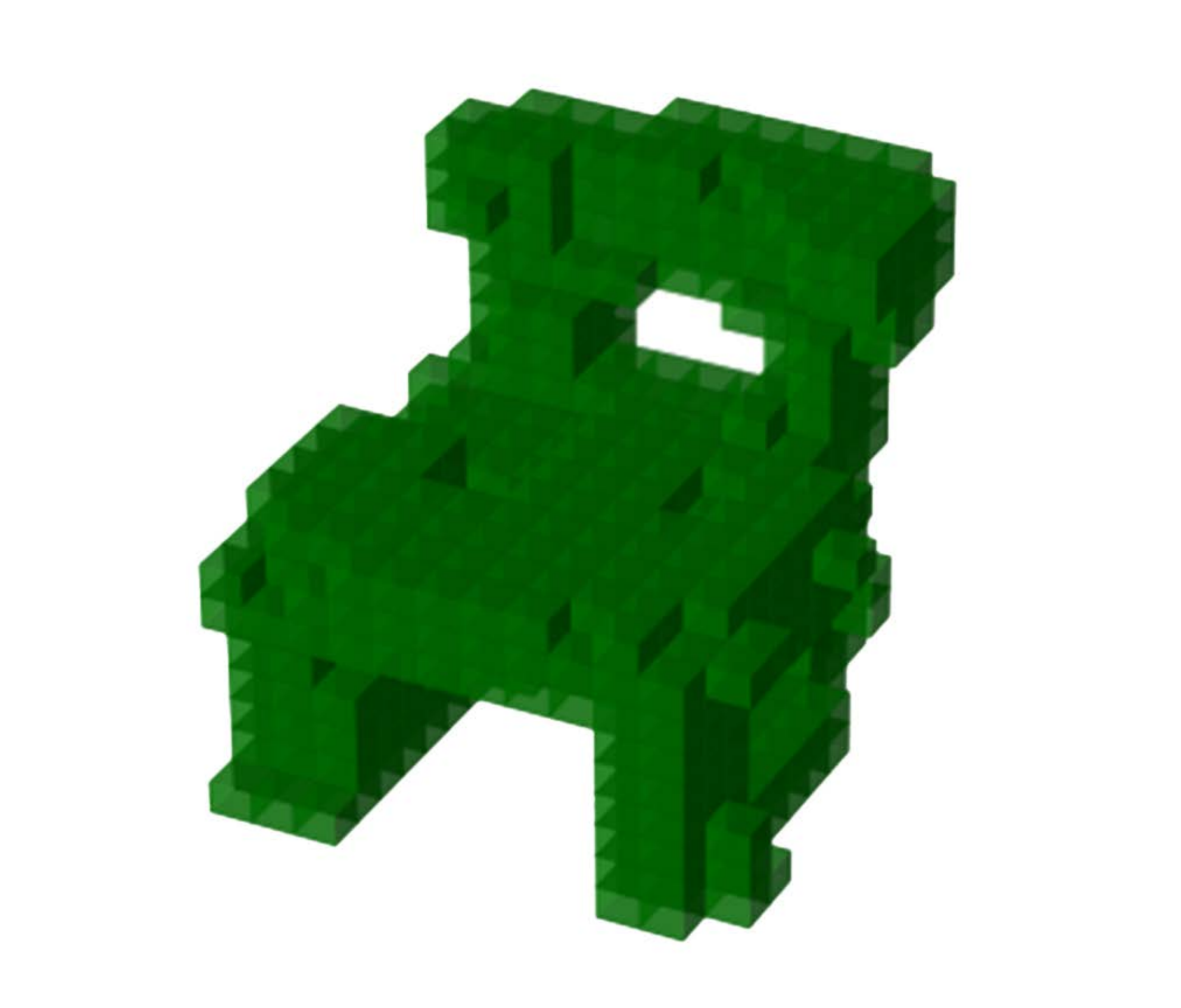}\label{fig:chair3}}\hfill
\subfigure[Chair 4.]{\includegraphics[width=0.24\linewidth]{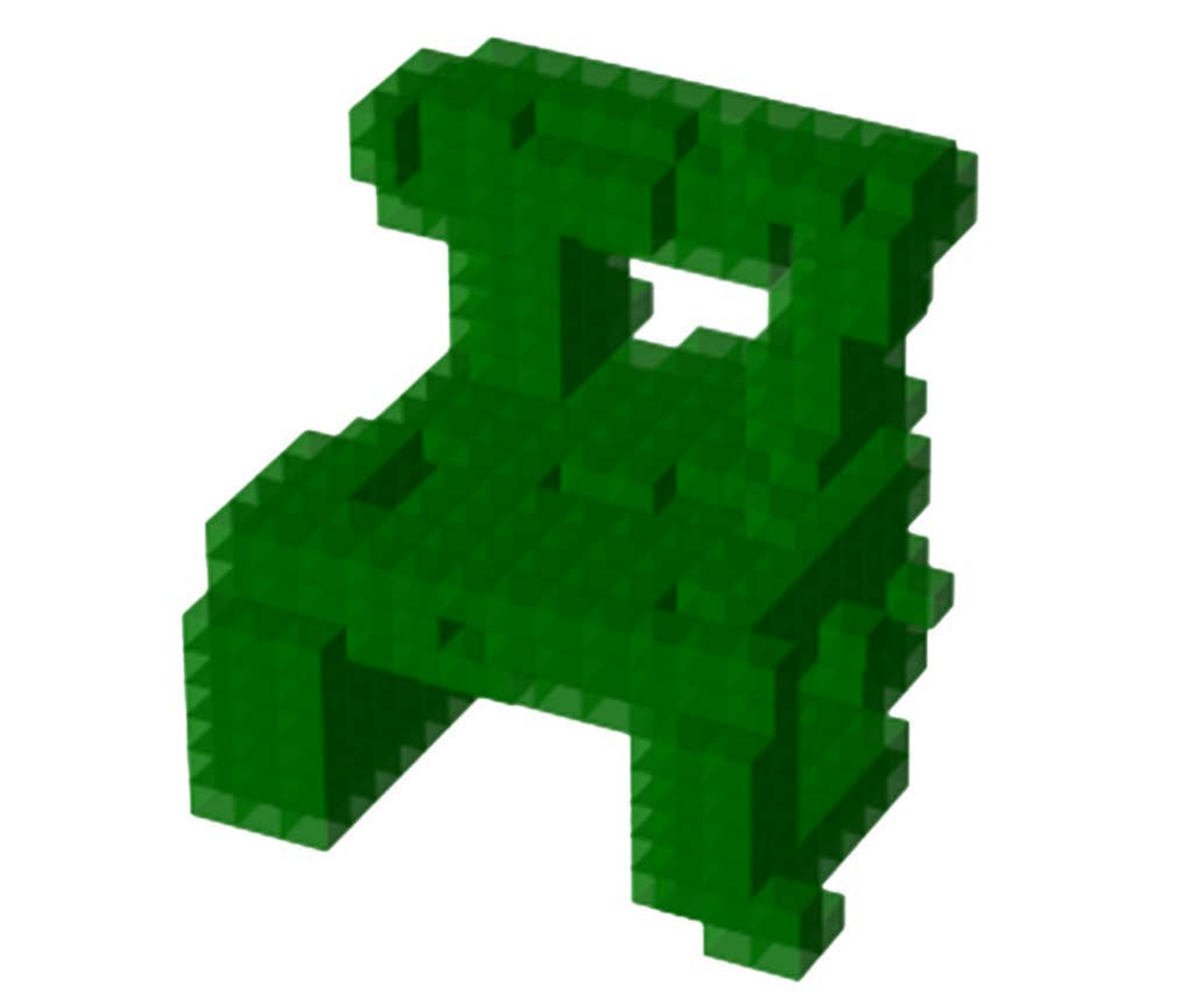}\label{fig:chair4}}\hfill

\vspace{-10pt}
    \caption{\footnotesize Comparisons of incomplete 3D Lego assemblies with various degrees of missing features with the same reference assembly. 
    Top: Incomplete assemblies with missing features. 
    Bottom: Finished assembly. 
    \label{fig:exp2}}
\end{figure}

In this part, we evaluate our framework's robustness in handling incomplete assemblies with varying degrees of missing features, as shown in \cref{fig:exp2}. 
Despite increasing amounts of key missing object features, the framework effectively reconstructs the final assemblies.

Starting with a missing top rail in \cref{fig:feat1}, the agent precisely reconstructs it. 
As more features are removed, the quality of the final assembly slightly decreases as reconstructing greater feature gaps with no exact end target becomes increasingly difficult. 
Regardless, the framework remains capable of producing high-similarity results while handling larger feature gaps and missing parts. 
The agent effectively demonstrates high robustness and accuracy.

\section{Conclusion and Future Work}
This paper proposes a novel formulation of the 3D assembly completion task.
Our approach combines PCD techniques with DRL to enable robots to infer assembly intent and optimize action sequencing.
We develop heuristics for action masking and a custom reward to assist agent training. 
Experiments validate the framework's capability in various construction scenarios, overcoming real-world constraints.
However, a limitation is that if the reference is significantly larger than the incomplete assembly, stability is not absolutely guaranteed throughout.
We aim to explore further optimization solutions to maximize stability in the future. 
In addition, we aim to incorporate temporary and permanent removal operations to allow for greater flexibility and efficiency in handling pre-existing problematic features.

\bibliographystyle{ifacconf}
\bibliography{ifacconf}  


\end{document}